\documentclass{ieeetj}

\usepackage{cite}
\usepackage{amsmath,amssymb,amsfonts}
\usepackage{graphicx,color}
\usepackage{xcolor}
\usepackage{textcomp}

\usepackage{tabularx}
\usepackage{booktabs}
\usepackage{multirow}
\usepackage{colortbl}
\usepackage[table]{xcolor}

\usepackage{float}
\usepackage{setspace}
\usepackage{pifont}
\usepackage{breakurl}
\usepackage[colorlinks=true, linkcolor=blue, citecolor=blue, urlcolor=black, breaklinks=true]{hyperref}
\usepackage{orcidlink}

\usepackage{tikz}
\usetikzlibrary{arrows.meta,positioning,fit,backgrounds,calc,shapes.geometric}
\usepackage{graphbox}

\definecolor{cFrozen}{RGB}{225,228,232}
\definecolor{cFrozenB}{RGB}{140,148,158}
\definecolor{cBasis}{RGB}{253,222,190}
\definecolor{cBasisB}{RGB}{214,137,42}
\definecolor{cLearn}{RGB}{205,239,217}
\definecolor{cLearnB}{RGB}{45,150,90}
\definecolor{cData}{RGB}{214,232,247}
\definecolor{cDataB}{RGB}{47,116,178}
\definecolor{cAlarm}{RGB}{250,219,216}
\definecolor{cAlarmB}{RGB}{192,57,43}

\tikzset{
  blk/.style   ={rectangle, rounded corners=2pt, draw=cFrozenB, fill=cFrozen,
                 minimum height=7mm, minimum width=12mm, align=center, font=\scriptsize},
  basis/.style ={rectangle, rounded corners=2pt, draw=cBasisB, fill=cBasis, thick,
                 minimum height=7mm, align=center, font=\scriptsize},
  learn/.style ={rectangle, rounded corners=2pt, draw=cLearnB, fill=cLearn, thick,
                 minimum height=7mm, align=center, font=\scriptsize},
  data/.style  ={rectangle, rounded corners=2pt, draw=cDataB, fill=cData,
                 minimum height=7mm, align=center, font=\scriptsize},
  alarm/.style ={rectangle, rounded corners=2pt, draw=cAlarmB, fill=cAlarm, thick,
                 minimum height=7mm, align=center, font=\scriptsize},
  op/.style    ={circle, draw=black!60, inner sep=0.7pt, font=\scriptsize},
  fl/.style    ={-{Latex[length=2mm]}, draw=black!65},
  fldash/.style={-{Latex[length=2mm]}, draw=black!50, dashed},
  lbl/.style   ={font=\scriptsize\itshape, text=black!65},
  note/.style  ={font=\scriptsize, align=center},
}

\def\BibTeX{{\rm B\kern-.05em{\sc i\kern-.025em b}\kern-.08em
    T\kern-.1667em\lower.7ex\hbox{E}\kern-.125emX}}
\AtBeginDocument{\definecolor{tmlcncolor}{cmyk}{0.93,0.59,0.15,0.02}\definecolor{NavyBlue}{RGB}{0,86,125}}

\def\OJlogo{\vspace{-4pt}$<$Society logo(s) and publication title will appear here.$>$}
\def\seclogo{\vspace{10pt}$<$Society logo(s) and publication title will appear here.$>$}

\def\authorrefmark#1{\ensuremath{^{\textbf{#1}}}}

\begin{document}
\receiveddate{XX Month, XXXX}
\reviseddate{XX Month, XXXX}
\accepteddate{XX Month, XXXX}
\publisheddate{XX Month, XXXX}
\currentdate{XX Month, XXXX}
\doiinfo{XXXX.2022.1234567}

\markboth{}{Author {et al.}}

\title{Transferable Low-Rank Convolutional Bases for Onboarding Unseen Medical Imaging Modalities}

\author{Ranat Das Prangon\authorrefmark{1}\orcidlink{0009-0001-5593-8384}, 
Istiaque Ahmed\authorrefmark{2}\orcidlink{0000-0002-3112-6568},
Shajid Hasan Naim\authorrefmark{3}\orcidlink{0009-0000-0901-4861},
\\ Waseem Mustak Zisan\authorrefmark{4}\orcidlink{0009-0008-4919-5255},
and Hossain Md Shakhawat\authorrefmark{5}\orcidlink{0000-0003-0713-0740}, Member, IEEE}
\affil{ Dept. of Chemical Engineering, Bangladesh University of Engineering and Technology (BUET), Dhaka, Bangladesh.}
\affil{ Graduate School of Informatics, Osaka Metropolitan University, Osaka, Japan.}
\affil{ Dept. of Mechanical Engineering, Chittagong University of Engineering and Technology (CUET), Chattogram, Bangladesh.}
\affil{ Dept. of Computer Science and Engineering, Bangladesh University of Engineering and Technology (BUET), Dhaka, Bangladesh.}
\affil{ School of Informatics, Kochi University of Technology, Kami, Kochi 782-8502, Japan.}
\corresp{Corresponding author: Istiaque Ahmed (email: sw23837u@st.omu.ac.jp) and Hossain Md Shakhawat (email: md.shakhawat@kochi-tech.ac.jp).}
\authornote{-All authors contributed equally to this work.}

\begin{abstract}
Deploying a medical imaging model that must later accommodate a modality it has never seen is a recurring practical problem: retraining the shared representation is expensive and destroys performance on the modalities already in service. We study this \emph{onboarding} problem under a strict leave-one-domain-out protocol, in which a convolutional backbone is pre-trained on source modalities (Kidney CT and Brain MRI), frozen permanently, and then required to accommodate an unseen modality (Chest X-ray). Under this protocol we establish three findings. First, decision-layer parameter-efficient fine-tuning is insufficient when the backbone has never observed the target modality: a linear probe and fully-connected LoRA both fall well short, whereas convolutional LoRA recovers most of the achievable accuracy, showing that adaptation must reach the convolutional features. Second, and centrally, the low-rank convolutional \emph{basis} learned on the source modalities \emph{transfers}: freezing that basis and training only its up-projections onboards the unseen modality using just $0.78\%$ of full fine-tuning's parameters, at an accuracy $6.11$ percentage points above a random basis of identical size, while an equivalent decision-layer basis exhibits no reliable transfer. Third, adapter-based onboarding leaves source-modality accuracy exactly unchanged ($\Delta = 0.00$ pp), whereas full fine-tuning reaches the highest target accuracy only by catastrophically degrading the source modalities. A Mahalanobis score on frozen backbone features detects the unseen modality with high sensitivity at a strict source-retention threshold, providing a practical trigger for when onboarding is required. All results are reported over three seeds with paired bootstrap confidence intervals.
\end{abstract}



\maketitle


\section{INTRODUCTION}
\IEEEpubidadjcol
\label{sec:introduction}

\IEEEPARstart{A}{clinical} imaging system rarely stays fixed. A model deployed for kidney CT and brain MRI may later be required to support chest radiography, and the modality that arrives was, by construction, not available when the shared representation was learned. This is the \emph{onboarding} problem: a new imaging modality must be accommodated by a system whose representation is already in service, without retraining that representation and without disturbing the modalities that depend on it \cite{Parisi2019,Guan2022}.

The obvious solution, fine-tuning the whole network on the new modality, is also the worst one. It is expensive, it produces a separate full model per modality, and it overwrites the representation that the existing modalities rely upon, causing catastrophic forgetting \cite{Kirkpatrick2017,Zenke2017,French1999}. Parameter-efficient fine-tuning (PEFT), and Low-Rank Adaptation (LoRA) in particular, appears to offer a way out: freeze the backbone and learn a small low-rank update per domain \cite{Hu2021,Dutt2024}. Because the shared weights are never modified, forgetting is avoided by construction.

There is, however, a methodological trap in how such systems are usually evaluated, and it is a trap we fell into ourselves. If the shared backbone is pre-trained on \emph{all} of the domains that will later be adapted, its frozen features are already close to linearly separable for every one of them. In that regime, a per-domain linear classifier matches LoRA, a domain router trained on visually distinct modalities is trivially accurate, and confidence calibration has nothing to correct. The experimental setting has quietly removed the very difficulties the method exists to address, and the resulting numbers say nothing about the deployment scenario that motivates the work. The realistic scenario is the opposite: the backbone has \emph{never} seen the modality being onboarded.

We therefore adopt a strict \emph{leave-one-domain-out} protocol. A convolutional backbone is pre-trained on source modalities only (Kidney CT and Brain MRI), frozen permanently, and then required to accommodate a modality withheld from pre-training entirely (Chest X-ray). The frozen features are now genuinely mismatched for the target, adaptation has real work to do, and the question of \emph{where} in the network that adaptation must occur becomes empirically decidable rather than assumed.

Within this protocol we ask a question that, to our knowledge, has not been posed for medical imaging PEFT: when a low-rank adaptation module is factorized into a down-projection basis $A$ and an up-projection $B$, does the basis $A$ learned on the source modalities carry information that \emph{transfers} to an unseen modality? If it does, a new modality can be onboarded by freezing $A$ and learning only $B$, at a small fraction of an adapter's cost. Because a low-rank bottleneck alone might account for any such benefit, every basis-transfer configuration is paired with a \emph{random-basis control} of identical parameter count, which isolates whether the \emph{learned} basis carries anything beyond the bottleneck structure.

Our contributions are as follows.

\begin{itemize}
    \item We formalize onboarding an unseen imaging modality onto a frozen backbone as a leave-one-domain-out problem, and show that conventional all-domains pre-training conceals the difficulty it is meant to measure.
    \item We show that decision-layer adaptation is insufficient here: a linear probe and FC-LoRA both fall well short, while convolutional LoRA recovers most of the achievable accuracy. Adaptation must reach the convolutional features.
    \item We show that the low-rank \emph{convolutional} basis learned on the source modalities transfers to an unseen one. Freezing it and training only the up-projections onboards at $0.78\%$ of full fine-tuning's cost, exceeding an identically sized random basis by $6.11$ pp. No equivalent transfer occurs at the decision layer.
    \item We convert ``zero forgetting'' from a tautology into a measurement: adapter onboarding leaves source accuracy exactly unchanged ($\Delta = 0.00$ pp), whereas full fine-tuning buys its target accuracy by degrading the source modalities substantially.
    \item We provide a deployable onboarding trigger: a Mahalanobis score on frozen features identifies the unseen modality far more reliably than max-softmax or energy scores at the same operating point.
\end{itemize}

All results are reported over three random seeds with paired bootstrap confidence intervals, and every comparison against the no-adaptation baseline is accompanied by an explicit significance test.

The remainder of this paper is organized as follows. Section~\ref{sec:Related_Work} reviews related work and positions our contribution against the mixture-of-experts LoRA and shared-basis LoRA literature. Section~\ref{sec:Methodology} presents the leave-one-domain-out protocol, the convolutional adaptation modules, and the basis-transfer mechanism. Section~\ref{sec:Experimental_Setup} describes the datasets, splits, and evaluation methodology. Section~\ref{sec:result_discussion} reports the results, and Section~\ref{sec:Conclusion} concludes.

\begin{table*}[t]
    \centering
    \caption{Positioning against representative parameter-efficient and multi-domain methods. We claim neither routed low-rank experts nor shared low-rank bases as novel; the distinguishing elements of this work are the leave-one-domain-out onboarding protocol, the localisation of necessary adaptation to the convolutional stage, and the matched random-basis control that establishes whether a \emph{learned} basis transfers across modalities.}
    \label{tab:related_work_comparison}
    \small
    \setlength{\tabcolsep}{4pt}
    \begin{tabular}{p{2.9cm} p{2.5cm} p{2.6cm} p{2.4cm} p{2.3cm} p{2.4cm}}
        \hline
        \rowcolor{gray!5}
        \textbf{Method} & \textbf{Adaptation site} & \textbf{Basis shared across domains} & \textbf{Modifies shared weights} & \textbf{Random-basis control} & \textbf{Onboards a modality unseen in pre-training} \\ \hline
        Full fine-tuning              & All weights          & ---                    & Yes & No  & Yes (with forgetting) \\
        EWC \cite{Kirkpatrick2017}    & All weights          & ---                    & Yes & No  & No \\
        Adapters \cite{Houlsby2019}   & Bottleneck (FC)      & No                     & No  & No  & No \\
        AdapterFusion \cite{Pfeiffer2020} & Bottleneck (FC)  & Composed by attention  & No  & No  & No \\
        LoRA \cite{Hu2021}            & Linear layers        & No                     & No  & No  & No \\
        MOELoRA \cite{Liu2024MOELoRA} & Linear layers        & No (routed experts)    & No  & No  & No \\
        TT-LoRA MoE \cite{Kunwar2025TTLoRA} & Linear layers  & No (routed experts)    & No  & No  & No \\
        VeRA \cite{Kopiczko2024VeRA}  & Linear layers        & Yes, \emph{random} and frozen & No & Is the method & No \\
        Tied-LoRA \cite{Renduchintala2024TiedLoRA} & Linear layers & Yes, tied across layers & No & No & No \\
        \textbf{This work}            & \textbf{Convolutional} & \textbf{Yes, \emph{learned} on source modalities and frozen} & \textbf{No} & \textbf{Yes, matched budget} & \textbf{Yes} \\ \hline
    \end{tabular}
\end{table*}

\section{RELATED WORK}
\label{sec:Related_Work}

Our contribution sits at the intersection of four lines of work: parameter-efficient adaptation, forgetting-aware modular learning, mixture-of-experts LoRA, and shared low-rank bases. We review each in turn and then state precisely what distinguishes the present study.

Parameter-efficient fine-tuning (PEFT) adapts a pre-trained model to a new task while updating only a small fraction of its parameters \cite{Houlsby2019,Dutt2024}. LoRA is the most widely adopted instance: a weight update is constrained to a low-rank product $\Delta W = AB$ while the base weights remain frozen \cite{Hu2021}. Subsequent work has varied the rank allocation \cite{Zhang2023}, quantized the frozen base \cite{dettmers2023qlora}, and applied low-rank updates at different depths \cite{Liu2021,He2022}. In medical imaging, PEFT has been shown to be competitive with full fine-tuning at a fraction of the cost \cite{Dutt2024,Wu2024}. Almost all of this literature applies low-rank updates to the linear or attention projections of transformer architectures. The question of \emph{where} adaptation must occur when the frozen representation is genuinely mismatched for the target domain is rarely posed, and it is central here.

A second line of work asks what happens to previously learned tasks when a model is adapted. Regularization methods such as Elastic Weight Consolidation and Learning without Forgetting penalize movement of parameters deemed important to earlier tasks \cite{Kirkpatrick2017,Li2017,Aljundi2017}, while modular and progressive architectures instead allocate task-specific capacity \cite{Rusu2016,Mallya2018}. Adapter-based methods insert lightweight per-task modules into a shared frozen model \cite{Houlsby2019,Bapna2019,Wang2020}, and AdapterFusion composes several such adapters through a learned attention mechanism \cite{Pfeiffer2020,Pfeiffer2021}. Methods in this family avoid forgetting \emph{by construction}, since the shared weights are never written to. We emphasize this explicitly rather than presenting it as an algorithmic achievement, and we make the corresponding claim falsifiable by measuring source-domain accuracy before and after onboarding against a full fine-tuning baseline that genuinely does forget.

Closer to our architecture, a substantial body of concurrent work combines LoRA experts with a routing mechanism. MOELoRA applies task-motivated LoRA experts with a learned gate for multi-task medical applications \cite{Liu2024MOELoRA}, TT-LoRA MoE decouples expert training from routing \cite{Kunwar2025TTLoRA}, and mixture-of-experts adapters have been used to improve continual learning of vision-language models \cite{Yu2024}. A frozen backbone with per-domain LoRA experts selected by a top-1 router is, structurally, a supervised mixture of experts, and we do not claim that architecture as a contribution. Our contribution is orthogonal to the routing question: we ask what must be adapted, and whether the \emph{factorized basis} of the adaptation itself is reusable across modalities. Routing enters this work only as an onboarding \emph{trigger} (Section~\ref{subsec:ood}), that is, as a mechanism for detecting that a modality outside the current repertoire has arrived.

Sharing structure across low-rank adapters is likewise not new. VeRA freezes a pair of randomly initialized projection matrices shared across all layers and learns only small scaling vectors \cite{Kopiczko2024VeRA}, Tied-LoRA ties the projection matrices across layers to reduce the parameter count \cite{Renduchintala2024TiedLoRA}, and LoRA-Hub composes previously trained LoRA modules for new tasks \cite{Huang2024LoraHub}. These methods share bases \emph{across layers or tasks within a single data distribution}, and, notably, VeRA's result that a \emph{random} frozen basis suffices is evidence that a low-rank bottleneck may carry benefit independently of what the basis contains.

That last observation is what makes the random-basis control indispensable, and it is precisely the comparison that defines our claim. We ask whether a basis \emph{learned on one set of imaging modalities} carries information that transfers to a modality \emph{never seen during pre-training}, and we test it against an identically sized random basis. Our finding is asymmetric and, to our knowledge, new: at the convolutional stage the learned basis outperforms a random one by $6.11$ pp, whereas at the decision layer it does not transfer reliably. Transferable low-rank structure therefore exists, but it resides in the convolutional feature adaptation rather than in the classifier.

Table~\ref{tab:related_work_comparison} contrasts the present work with representative alternatives along these dimensions. We claim neither routed LoRA experts nor shared low-rank bases as novel. The contributions are (i) the leave-one-domain-out onboarding protocol, which exposes a difficulty that all-domains pre-training conceals; (ii) the empirical localization of necessary adaptation to the convolutional stage under that protocol; and (iii) the demonstration, with matched-budget random-basis controls, that a convolutional low-rank basis learned on source modalities transfers to an unseen one, enabling onboarding at $0.78\%$ of the cost of full fine-tuning with exactly zero forgetting.

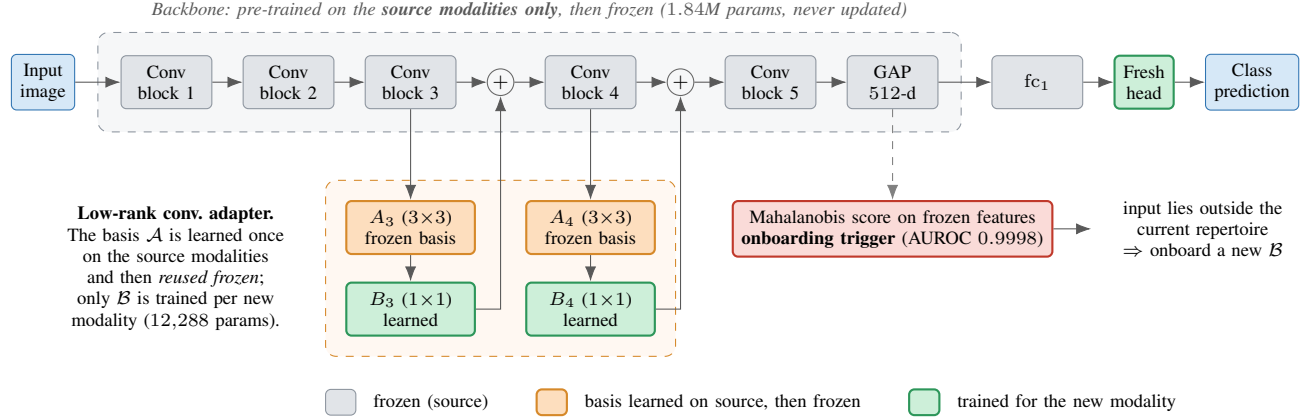
\begin{figure*}[!t]
\centering
\resizebox{0.98\textwidth}{!}{
\begin{tikzpicture}[node distance=4mm]

\node[data] (img) {Input\\image};
\node[blk, right=6mm of img] (c1) {Conv\\block 1};
\node[blk, right=of c1]      (c2) {Conv\\block 2};
\node[blk, right=of c2]      (c3) {Conv\\block 3};
\node[op,  right=of c3]      (p3) {$+$};
\node[blk, right=of p3]      (c4) {Conv\\block 4};
\node[op,  right=of c4]      (p4) {$+$};
\node[blk, right=of p4]      (c5) {Conv\\block 5};
\node[blk, right=of c5]      (gap){GAP\\$512$-d};

\draw[fl] (img)--(c1); \draw[fl] (c1)--(c2); \draw[fl] (c2)--(c3);
\draw[fl] (c3)--(p3);  \draw[fl] (p3)--(c4); \draw[fl] (c4)--(p4);
\draw[fl] (p4)--(c5);  \draw[fl] (c5)--(gap);

\begin{scope}[on background layer]
  \node[draw=cFrozenB, dashed, rounded corners, fill=cFrozen!30,
        fit=(c1)(gap), inner sep=3mm] (bb) {};
\end{scope}
\node[lbl, above=0.5mm of bb.north]
     {Backbone: pre-trained on the \textbf{source modalities only}, then frozen ($1.84$M params, never updated)};

\node[blk,   right=7mm of gap] (fc1)  {$\mathrm{fc}_1$};
\node[learn, right=of fc1]     (head) {Fresh\\head};
\node[data,  right=of head]    (pred) {Class\\prediction};
\draw[fl] (gap)--(fc1); \draw[fl] (fc1)--(head); \draw[fl] (head)--(pred);

\node[basis, below=12mm of c3, minimum width=17mm] (A3) {$A_3$ ($3{\times}3$)\\frozen basis};
\node[learn, below=3.5mm of A3, minimum width=17mm] (B3) {$B_3$ ($1{\times}1$)\\learned};
\node[basis, below=12mm of c4, minimum width=17mm] (A4) {$A_4$ ($3{\times}3$)\\frozen basis};
\node[learn, below=3.5mm of A4, minimum width=17mm] (B4) {$B_4$ ($1{\times}1$)\\learned};

\draw[fl] (c3.south) -- (A3.north);      
\draw[fl] (A3) -- (B3);
\draw[fl] (B3.east) -| (p3.south);       
\draw[fl] (c4.south) -- (A4.north);      
\draw[fl] (A4) -- (B4);
\draw[fl] (B4.east) -| (p4.south);       

\begin{scope}[on background layer]
  \node[draw=cBasisB, dashed, rounded corners, fill=cBasis!20,
        fit=(A3)(B3)(A4)(B4), inner sep=2.6mm] (ad) {};
\end{scope}
\node[note, left=4mm of ad.west, text width=29mm, anchor=east]
     {\textbf{Low-rank conv.\ adapter.}\\
      The basis $\mathcal{A}$ is learned once on the source modalities and then
      \emph{reused frozen}; only $\mathcal{B}$ is trained per new modality
      ($12{,}288$ params).};

\node[alarm, below=12mm of gap, minimum width=32mm, anchor=north] (ood)
     {Mahalanobis score on frozen features\\
      \textbf{onboarding trigger}\ (AUROC $0.9998$)};
\draw[fldash] (gap.south) -- (ood.north);
\node[note, right=5mm of ood.east, text width=27mm, anchor=west] (oodnote)
     {input lies outside the current repertoire\\$\Rightarrow$ onboard a new $\mathcal{B}$};
\draw[fl] (ood.east) -- (oodnote.west);

\coordinate (lg) at ($(ad.south west)+(0,-6mm)$);
\node[blk,   anchor=west, minimum width=4mm, minimum height=3.4mm] (l1) at (lg) {};
\node[note,  right=1mm of l1, anchor=west] (t1) {frozen (source)};
\node[basis, right=5mm of t1.east, minimum width=4mm, minimum height=3.4mm] (l2) {};
\node[note,  right=1mm of l2, anchor=west] (t2) {basis learned on source, then frozen};
\node[learn, right=5mm of t2.east, minimum width=4mm, minimum height=3.4mm] (l3) {};
\node[note,  right=1mm of l3, anchor=west] (t3) {trained for the new modality};

\end{tikzpicture}}
\caption{System architecture. A convolutional backbone is pre-trained on the source modalities only and then frozen permanently (grey). An unseen modality is onboarded by attaching a low-rank convolutional adapter, whose down-projection basis $\mathcal{A}$ may be reused from the source modalities (orange, frozen) so that only the up-projections $\mathcal{B}$ and a fresh classification head are trained (green). A Mahalanobis score computed on the frozen features acts as the onboarding trigger, signalling that an input lies outside the current modality repertoire.}
\label{fig:figure1}
\end{figure*}

\section{METHODOLOGY}
\label{sec:Methodology}

We first state the onboarding problem and the leave-one-domain-out protocol (Section~\ref{subsec:protocol}). We then define the adaptation modules under comparison, which differ only in \emph{where} the low-rank update is applied and in \emph{whether} its basis is learned, transferred, or random (Sections~\ref{subsec:conv_lora}--\ref{subsec:basis}). Finally we describe the onboarding trigger (Section~\ref{subsec:ood}).

\subsection{PROBLEM FORMULATION AND LEAVE-ONE-DOMAIN-OUT PROTOCOL}
\label{subsec:protocol}

Let $\mathcal{D}_{\text{src}} = \{\mathcal{D}_1, \dots, \mathcal{D}_k\}$ be a set of source imaging modalities available when the shared representation is trained, and let $\mathcal{D}_{\text{tgt}}$ be a modality that becomes available only afterwards. A backbone $\phi$ is trained on $\bigcup_{i} \mathcal{D}_i$ and then frozen. The onboarding problem is to attain high accuracy on $\mathcal{D}_{\text{tgt}}$ by training a small set of new parameters $\theta_{\text{tgt}}$, subject to two constraints: no parameter shared with $\mathcal{D}_{\text{src}}$ may be modified, and $|\theta_{\text{tgt}}| \ll |\phi|$.

The essential methodological point is that $\mathcal{D}_{\text{tgt}}$ must be withheld from the pre-training of $\phi$. If $\phi$ is trained on all modalities, its frozen features are close to linearly separable for each of them, and a per-domain linear head suffices; the resulting comparison is uninformative about deployment, where the arriving modality is by definition absent from pre-training. We therefore instantiate the protocol with $\mathcal{D}_{\text{src}} = \{\text{Kidney CT}, \text{Brain MRI}\}$ and $\mathcal{D}_{\text{tgt}} = \text{Chest X-ray}$, and no chest radiograph is used at any point before onboarding.

The backbone is a five-block convolutional network with channel widths $3 \!\to\! 32 \!\to\! 64 \!\to\! 128 \!\to\! 256 \!\to\! 512$, each block comprising a $3\!\times\!3$ convolution, batch normalization, and ReLU, with max-pooling after the first four blocks and global average pooling after the fifth, followed by a fully connected layer $\mathrm{fc}_1: \mathbb{R}^{512} \!\to\! \mathbb{R}^{512}$. It contains $1.84 \times 10^{6}$ parameters. Writing $\mathbf{f} = \phi(x) \in \mathbb{R}^{512}$ for the pooled features, every onboarding variant computes
\begin{equation}
\mathbf{h} = \mathrm{ReLU}\!\left( \mathrm{fc}_1(\mathbf{f}) + \Delta_{\text{fc}}(\mathbf{f}) \right), \qquad
\mathbf{z} = \mathbf{W}_{\text{head}}\,\mathbf{h} + \mathbf{b}_{\text{head}},
\label{eq:unified_forward}
\end{equation}
where $\Delta_{\text{fc}}$ is an optional decision-layer low-rank update and $\mathbf{W}_{\text{head}} \in \mathbb{R}^{4 \times 512}$ is a freshly initialized classification head for the target modality. Convolutional adaptation, where present, modifies $\mathbf{f}$ itself, as defined next.

\textbf{Fairness of comparison.} Every variant in this study receives an \emph{identical, freshly initialized} head. This matters: if only the no-adaptation baseline is given a fresh head while the low-rank variants must additionally repair an untrained output block through a frozen basis, the comparison is confounded and penalizes basis transfer. Under Eq.~\ref{eq:unified_forward}, the sole difference between variants is the adaptation mechanism.

\subsection{CONVOLUTIONAL LOW-RANK ADAPTATION}
\label{subsec:conv_lora}

Standard LoRA restricted to fully connected layers can only reshape the decision boundary over features that the frozen backbone already produces. When the backbone has never observed the target modality, those features are mismatched, and no reweighting of them recovers the information the convolutional stages failed to extract. We therefore inject low-rank updates into the convolutional stages themselves.

For a convolutional block $i$ with $C^{\text{in}}_i$ input and $C^{\text{out}}_i$ output channels, let $\mathbf{x}_i$ denote its input activation. We define the low-rank convolutional update
\begin{equation}
\Delta_i(\mathbf{x}_i) \;=\; \frac{\alpha}{r}\, B_i\!\left( A_i(\mathbf{x}_i) \right),
\label{eq:conv_lora}
\end{equation}
where $A_i$ is a $3\!\times\!3$ convolution mapping $C^{\text{in}}_i \!\to\! r$ channels (the \emph{down-projection basis}), $B_i$ is a $1\!\times\!1$ convolution mapping $r \!\to\! C^{\text{out}}_i$ channels (the \emph{up-projection}), $r \ll \min(C^{\text{in}}_i, C^{\text{out}}_i)$ is the rank, and $\alpha$ is a scaling factor. The update is added to the raw convolution output before normalization and activation:
\begin{equation}
\mathbf{x}_{i+1} \;=\; \mathrm{Pool}\Big( \mathrm{ReLU}\big( \mathrm{BN}_i\big( W_i * \mathbf{x}_i + \Delta_i(\mathbf{x}_i) \big) \big) \Big).
\label{eq:conv_block}
\end{equation}
Each $B_i$ is initialized to zero, so $\Delta_i \equiv 0$ at initialization and the adapted network is exactly the frozen backbone; the base weights $W_i$ are never updated. The set of adapted blocks $\mathcal{L}$ and the rank $r$ jointly determine the adaptation capacity, and are swept in Section~\ref{subsec:capacity}.

\subsection{BASIS TRANSFER AND THE RANDOM-BASIS CONTROL}
\label{subsec:basis}

Equation~\ref{eq:conv_lora} factorizes the adaptation into a basis $A_i$, which selects a low-dimensional subspace of the input activations, and an up-projection $B_i$, which maps that subspace onto the output channels. This raises the question at the heart of this paper: is $A_i$ modality-specific, or does it capture adaptation directions that are common to medical imaging modalities in general?

\textbf{Learning a shared basis on the source modalities.} During source training we learn a single basis $\mathcal{A} = \{A_i\}_{i \in \mathcal{L}}$ \emph{jointly} across all source modalities, while each source modality $d$ retains its own up-projections $\mathcal{B}^{(d)}$ and head. Batches from the source modalities are interleaved within each epoch, so that
\begin{equation}
\mathcal{A}^{\star} = \arg\min_{\mathcal{A}} \; \sum_{d \in \mathcal{D}_{\text{src}}} \; \min_{\mathcal{B}^{(d)},\, \mathbf{W}^{(d)}_{\text{head}}} \; \mathcal{L}_{\text{CE}}\big(d;\, \mathcal{A}, \mathcal{B}^{(d)}, \mathbf{W}^{(d)}_{\text{head}}\big).
\label{eq:shared_basis}
\end{equation}
Interleaving is essential: were the modalities presented sequentially, the last would overwrite the shared directions learned for the others, reintroducing precisely the interference the design seeks to avoid.

\textbf{Onboarding by basis transfer.} To onboard $\mathcal{D}_{\text{tgt}}$, the basis $\mathcal{A}^{\star}$ is \emph{frozen} and reused, and only the target up-projections $\mathcal{B}^{(\text{tgt})}$ and a fresh head are trained. Because $\mathcal{A}^{\star}$ is shared and never rewritten, and the backbone is frozen, no parameter on which the source modalities depend is modified; forgetting is therefore exactly zero, a claim we verify empirically in Section~\ref{subsec:forgetting}.

The parameter cost is markedly asymmetric. A full convolutional adapter costs $\sum_{i \in \mathcal{L}} (9\, r\, C^{\text{in}}_i + r\, C^{\text{out}}_i)$ parameters, of which the $3\!\times\!3$ basis dominates; basis transfer pays only the $1\!\times\!1$ up-projections, $\sum_{i \in \mathcal{L}} r\, C^{\text{out}}_i$. For $\mathcal{L} = \{3,4\}$ and $r=16$ this reduces the per-modality cost from $67{,}584$ to $12{,}288$ parameters.

\textbf{The random-basis control.} A low-rank bottleneck may confer benefit regardless of what its basis contains; indeed, VeRA \cite{Kopiczko2024VeRA} shows that a frozen \emph{random} basis can suffice in some settings. Any claim that a \emph{learned} basis transfers is therefore meaningless without a matched control. For every basis-transfer configuration we train an otherwise identical variant in which $\mathcal{A}$ is randomly initialized and frozen, with exactly the same number of trainable parameters. The contrast between the two isolates the contribution of the \emph{learned} basis from that of the bottleneck structure, and it is this contrast, rather than accuracy alone, that we report as the central result.

\begin{figure}[!t]
\centering
\resizebox{\linewidth}{!}{
\begin{tikzpicture}[node distance=4mm]

\node[data] (x) {$\mathbf{x}_i$};
\node[blk, right=10mm of x, minimum width=21mm] (W) {frozen conv $W_i$\\($3{\times}3$, never updated)};
\node[basis, below=8mm of W, minimum width=21mm] (A) {down-projection $A_i$\\$3{\times}3$,\; $C^{\mathrm{in}}_i\!\to\! r$};
\node[learn, below=3.5mm of A, minimum width=21mm] (B) {up-projection $B_i$\\$1{\times}1$,\; $r\!\to\! C^{\mathrm{out}}_i$,\; zero-init};
\node[learn, below=3.5mm of B, minimum width=21mm] (S) {scale $\alpha/r$};

\node[op,  right=8mm of W] (plus) {$+$};
\node[blk, right=5mm of plus, minimum width=15mm] (bn) {BN $\to$ ReLU\\$\to$ pool};
\node[data, right=5mm of bn] (out) {$\mathbf{x}_{i+1}$};

\draw[fl] (x.east) -- (W.west);
\draw[fl] (x.south) |- (A.west);
\draw[fl] (A) -- (B);
\draw[fl] (B) -- (S);
\draw[fl] (W.east) -- (plus.west);
\draw[fl] (S.east) -| (plus.south);
\draw[fl] (plus.east) -- (bn.west);
\draw[fl] (bn.east) -- (out.west);

\node[lbl, above=1mm of W.north] {frozen path};
\node[lbl, left=1.5mm of A.west, anchor=east, text width=12mm, align=right] {adaptation\\path};

\node[note, below=4mm of S.south, text width=70mm] (eq)
     {$\Delta_i(\mathbf{x}_i)=\tfrac{\alpha}{r}\,B_i\!\left(A_i(\mathbf{x}_i)\right)$\ \ added to the raw
      convolution output. As $B_i$ is zero-initialised, $\Delta_i\!\equiv\!0$ initially and the
      adapted network is exactly the frozen backbone.};

\node[note, below=6mm of eq.south west, anchor=north west, font=\scriptsize\bfseries] (hdr)
     {The three conditions on $A_i$, compared at \emph{identical} trainable budget:};

\node[basis, below=3mm of hdr.west, anchor=north west, minimum width=27mm, minimum height=8.5mm] (v1)
     {$A_i$ \textbf{learned} on the new modality};
\node[note, right=2.5mm of v1, anchor=west, text width=42mm] (n1)
     {\textsc{Conv-LoRA} --- a full adapter; both $A_i$ and $B_i$ are trained. $69{,}636$ params.};

\node[basis, below=3mm of v1.south west, anchor=north west, minimum width=27mm, minimum height=8.5mm] (v2)
     {$A_i$ \textbf{reused frozen} from the source modalities};
\node[note, right=2.5mm of v2, anchor=west, text width=42mm] (n2)
     {\textsc{Conv basis-transfer} (\textbf{ours}) --- only $B_i$ is trained. $14{,}340$ params.};

\node[alarm, below=3mm of v2.south west, anchor=north west, minimum width=27mm, minimum height=8.5mm] (v3)
     {$A_i$ \textbf{random} and frozen};
\node[note, right=2.5mm of v3, anchor=west, text width=42mm] (n3)
     {\textsc{Random-basis control} --- identical budget; isolates whether the \emph{learned} basis carries anything beyond the bottleneck.};

\begin{scope}[on background layer]
  \node[draw=black!35, dashed, rounded corners,
        fit=(hdr)(v1)(v2)(v3)(n1)(n2)(n3), inner sep=2.8mm] {};
\end{scope}

\end{tikzpicture}}
\caption{Convolutional low-rank adaptation and basis transfer. The frozen path (grey) is the pre-trained convolution. The adaptation path factorizes into a $3\!\times\!3$ down-projection basis $A_i$ and a $1\!\times\!1$ up-projection $B_i$. In basis transfer, $A_i$ is inherited frozen from the source modalities (orange) and only $B_i$ is learned (green); the random-basis control replaces the learned $A_i$ with a frozen random one at identical cost.}
\label{fig:figure2}
\end{figure}

\begin{figure}[!t]
\centering
\resizebox{\linewidth}{!}{
\begin{tikzpicture}[node distance=4mm]

\node[note, anchor=west, font=\scriptsize\bfseries] (h1) at (0,0)
     {\textbf{1.}\ \ Pre-train on the source modalities, then freeze};
\node[data, below=3mm of h1.west, anchor=north west, minimum width=21mm] (src)
     {Kidney CT\ +\ Brain MRI\\\scriptsize (source modalities)};
\node[blk, right=6mm of src, minimum width=19mm] (bb) {Backbone $\phi$\\$1.84$M params};
\node[blk, right=5mm of bb, minimum width=13mm, draw=cFrozenB, very thick] (frz)
     {\textbf{FROZEN}\\forever};
\draw[fl] (src)--(bb); \draw[fl] (bb)--(frz);

\node[note, below=7mm of src.south west, anchor=north west, font=\scriptsize\bfseries] (h2)
     {\textbf{2.}\ \ Learn a shared low-rank basis on the source modalities};
\node[data, below=3mm of h2.west, anchor=north west, minimum width=21mm] (src2)
     {source batches\\\scriptsize interleaved, not sequential};
\node[basis, right=6mm of src2, minimum width=19mm] (A) {shared basis $\mathcal{A}$\\$55{,}296$ params};
\node[basis, right=5mm of A, minimum width=13mm, very thick] (frz2) {\textbf{FROZEN}\\reused later};
\draw[fl] (src2)--(A); \draw[fl] (A)--(frz2);
\node[note, below=2mm of src2.south west, anchor=north west, text width=76mm] (n2)
     {Interleaving matters: trained sequentially, the last modality would overwrite the shared directions learned for the others.};

\node[note, below=7mm of n2.south west, anchor=north west, font=\scriptsize\bfseries] (h3)
     {\textbf{3.}\ \ An \emph{unseen} modality arrives, and is detected};
\node[data, below=3mm of h3.west, anchor=north west, minimum width=21mm,
      draw=cAlarmB, fill=cAlarm] (tgt)
     {Chest X-ray\\\scriptsize never seen in steps 1--2};
\node[alarm, right=6mm of tgt, minimum width=26mm] (det)
     {Mahalanobis on frozen features\\$100\%$ detected @ $95\%$ TNR};
\node[note, right=3mm of det.east, anchor=west, text width=14mm] (n3) {$\Rightarrow$ onboard a new $\mathcal{B}$};
\draw[fl] (tgt)--(det); \draw[fl] (det)--(n3);

\node[note, below=7mm of tgt.south west, anchor=north west, font=\scriptsize\bfseries] (h4)
     {\textbf{4.}\ \ Onboard: train \emph{only} the up-projections and a fresh head};
\node[blk,   below=3mm of h4.west, anchor=north west, minimum width=16mm] (o1) {frozen $\phi$};
\node[basis, right=3mm of o1, minimum width=17mm] (o2) {frozen basis $\mathcal{A}$};
\node[learn, right=3mm of o2, minimum width=19mm] (o3) {\textbf{train} $\mathcal{B}$ + head\\$14{,}340$ params};
\node[data,  right=3mm of o3, minimum width=14mm] (o4) {$\mathbf{87.29\%}$\\\scriptsize new modality};
\draw[fl] (o1)--(o2); \draw[fl] (o2)--(o3); \draw[fl] (o3)--(o4);
\node[note, below=2.5mm of o1.south west, anchor=north west, text width=80mm] (n4)
     {Nothing the source modalities depend on is modified, so their accuracy is unchanged
      \emph{exactly} ($\Delta=0.00$ pp). Full fine-tuning reaches $94.11\%$ here, but degrades
      the source modalities by $65.39$ pp.};

\begin{scope}[on background layer]
  \node[draw=cFrozenB!70, rounded corners, fill=cFrozen!20, fit=(h1)(src)(frz), inner sep=2.4mm] {};
  \node[draw=cBasisB!70,  rounded corners, fill=cBasis!14,  fit=(h2)(src2)(frz2)(n2), inner sep=2.4mm] {};
  \node[draw=cAlarmB!70,  rounded corners, fill=cAlarm!14,  fit=(h3)(tgt)(det)(n3), inner sep=2.4mm] {};
  \node[draw=cLearnB!70,  rounded corners, fill=cLearn!14,  fit=(h4)(o1)(o4)(n4), inner sep=2.4mm] {};
\end{scope}

\end{tikzpicture}}
\caption{Leave-one-domain-out onboarding protocol. The backbone and the shared basis are learned from the source modalities only and then frozen; the target modality is withheld entirely until onboarding, at which point only the up-projections and a fresh head are trained.}
\label{fig:figure3}
\end{figure}
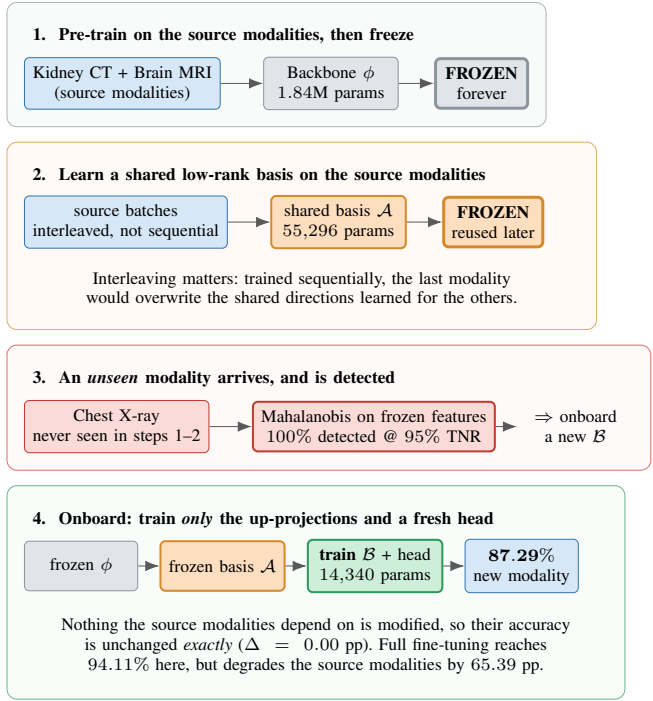

\subsection{ONBOARDING TRIGGER VIA OUT-OF-DISTRIBUTION DETECTION}
\label{subsec:ood}

Onboarding presupposes knowing that a new modality has arrived. We treat this as an out-of-distribution detection problem over the frozen backbone features, and we deliberately judge detectors at a deployment-relevant operating point rather than by ranking metrics alone.

Let $\pi$ be a lightweight router trained on the source modalities to predict modality identity from $\mathbf{f} = \phi(x)$, producing logits $\boldsymbol{\ell}(x) \in \mathbb{R}^{k}$. We compare three scores, where a higher value indicates a more likely unseen modality:
\begin{align}
s_{\text{msp}}(x) &= 1 - \max_j \; \mathrm{softmax}(\boldsymbol{\ell}(x))_j, \label{eq:msp}\\
s_{\text{energy}}(x) &= -\log \textstyle\sum_j \exp\big(\boldsymbol{\ell}(x)_j\big), \label{eq:energy}\\
s_{\text{maha}}(x) &= \min_{d \in \mathcal{D}_{\text{src}}} \; \big(\mathbf{f} - \boldsymbol{\mu}_d\big)^{\!\top} \boldsymbol{\Sigma}^{-1} \big(\mathbf{f} - \boldsymbol{\mu}_d\big), \label{eq:maha}
\end{align}
where $\boldsymbol{\mu}_d$ is the mean feature of source modality $d$ and $\boldsymbol{\Sigma}$ is the covariance shared across source modalities, both estimated on source \emph{training} data only.

We report AUROC but base our conclusions on the \emph{detection rate at $95\%$ source retention}: the fraction of target-modality inputs flagged when the threshold is set to admit $95\%$ of source inputs. A detector with high AUROC but low detection at this threshold is not a usable safety gate, and, as Section~\ref{subsec:ood_results} shows, the two criteria disagree sharply in practice.

\section{EXPERIMENTAL SETUP}
\label{sec:Experimental_Setup}

We evaluate on three publicly available medical imaging datasets, each with four diagnostic classes, spanning three distinct acquisition modalities. The CT-KIDNEY dataset \cite{Nazmul0087CTKidney} contains axial computed-tomography slices in four classes (Cyst, Normal, Stone, and Tumor) and the Brain Tumor MRI dataset \cite{NickparvarBrainTumorMRI} contains magnetic-resonance images in four classes (Glioma, Meningioma, No Tumor, and Pituitary); these two serve as the source modalities. The Chest X-ray dataset \cite{ChestXrayPCT} contains radiographs in four classes (COVID-19, Normal, Pneumonia, and Tuberculosis) and serves as the target: it is withheld from backbone pre-training and from shared-basis learning, and is introduced only at onboarding. To balance the modalities, each class is capped at $1{,}500$ images sampled without replacement. The three modalities occupy disjoint, contiguous blocks of a $12$-class global label space (Kidney $0$--$3$, Brain $4$--$7$, Chest $8$--$11$), so that domain identity and class identity remain separable throughout.

Because the validity of the protocol rests entirely on the target modality never influencing the frozen representation, we control the splits strictly. Each modality is split $60\%$ / $15\%$ / $25\%$ into training, validation, and test sets, stratified by class. The splits are disjoint, and disjointness is asserted programmatically at construction time rather than assumed. The validation split is used only for model selection and for estimating the covariance of the Mahalanobis detector; it is never used to fit adapters, bases, or heads. The test split is untouched until final evaluation. Where a source dataset ships pre-partitioned directories, images are de-duplicated by filename before re-splitting, so that the same image cannot appear in two partitions. We note explicitly that none of these public datasets exposes patient identifiers, so patient-level disjointness cannot be guaranteed; this is a limitation of the data rather than of the protocol, and we discuss its implications in Section~\ref{subsec:limitations}.

Turning to implementation, images are resized to $128 \times 128$, normalized per channel with $\mu = \sigma = 0.5$, and augmented during training with random horizontal flips ($p=0.5$). All models are optimized with Adam at a learning rate of $1 \times 10^{-3}$ and batch size $32$. The backbone is pre-trained for $12$ epochs on the source modalities; adapters, bases, and heads are trained for $12$ epochs; the router for $10$ epochs. Fully connected LoRA uses rank $r=64$, convolutional LoRA uses rank $r=16$ applied to blocks $\mathcal{L}=\{3,4\}$ unless stated otherwise, and the scaling factor is $\alpha = 32$ throughout. Experiments were run on an NVIDIA T4 GPU.

Nine onboarding variants are compared. All of them adapt the same frozen source-only backbone and receive an identical, freshly initialized target head (Eq.~\ref{eq:unified_forward}), so that the only difference between them is the adaptation mechanism. The \emph{linear probe} ($2{,}052$ trainable parameters) applies no adaptation at all and serves as the baseline. Three variants adapt at conventional sites: \emph{FC-LoRA} ($67{,}588$) places a low-rank update at the decision layer, \emph{Conv-LoRA} ($69{,}636$) places one at convolutional blocks $\{3,4\}$, and \emph{Conv+FC-LoRA} ($135{,}172$) applies both. The central comparisons are two matched pairs. At the decision layer, \emph{FC basis-transfer} ($34{,}820$) trains only the up-projection against a basis frozen from the source modalities, and \emph{FC random-basis} ($34{,}820$) is its control with a frozen random basis of identical size. At the convolutional stage, \emph{CONV basis-transfer} ($14{,}340$) and \emph{CONV random-basis} ($14{,}340$) form the corresponding pair. Finally, \emph{full fine-tuning} ($1{,}841{,}424$) updates all backbone parameters; it is the upper bound on target accuracy and the only variant that modifies shared parameters.

Evaluation follows four criteria. Target-modality classification accuracy is measured on the held-out test split and reported as the mean $\pm$ standard deviation over three random seeds ($42$, $1337$, $2024$), each inducing an independent split and an independent backbone. For every variant we report the difference against the linear probe with a $95\%$ confidence interval obtained by \emph{paired bootstrap} ($2{,}000$ resamples) over per-sample correctness, pooled across seeds with pairing preserved within each seed; a difference is called significant only when the interval excludes zero. We adopt this rather than comparing point estimates because, at these sample sizes, differences of one to two percentage points are not distinguishable from noise. Forgetting is quantified as source-modality test accuracy measured before and after onboarding, for both adapter-based onboarding and full fine-tuning. Detection is reported as AUROC together with the detection rate at $95\%$ source retention, for each of the three scores in Eqs.~\ref{eq:msp}--\ref{eq:maha}.

Parameter counts require care, and we report trainable parameters for the newly onboarded modality alongside the same figure expressed as a percentage of full fine-tuning; total stored and inference-active parameters are reported separately in Section~\ref{subsec:params}. Conflating these three quantities is a common source of overstated efficiency claims, and we keep them distinct throughout.
\section{EVALUATION AND DISCUSSION}
\label{sec:result_discussion}

We first establish that decision-layer adaptation is insufficient when the backbone has never seen the target modality (Section~\ref{subsec:onboarding}). We then present the central result, that the convolutional low-rank basis transfers across modalities while the decision-layer basis does not (Section~\ref{subsec:transfer}). We verify the zero-forgetting claim against a full fine-tuning baseline that genuinely forgets (Section~\ref{subsec:forgetting}), examine how much of the remaining gap convolutional capacity can close (Section~\ref{subsec:capacity}), and evaluate the onboarding trigger (Section~\ref{subsec:ood_results}).

\begin{table*}[t]
    \centering
    \caption{Onboarding the unseen modality (Chest X-ray) onto a backbone pre-trained on Kidney CT and Brain MRI only. All variants adapt the same frozen backbone and receive an identical freshly initialised head; they differ only in the adaptation mechanism. Accuracy is the mean $\pm$ standard deviation over three seeds. Differences are measured against the linear probe by paired bootstrap (2{,}000 resamples) over per-sample correctness pooled across seeds; a difference is marked significant only when the $95\%$ interval excludes zero. The two shaded pairs are the matched learned-versus-random basis comparisons.}
    \label{tab:onboarding}
    \small
    \setlength{\tabcolsep}{5pt}
    \begin{tabular}{p{4.3cm} r r p{2.6cm} r p{2.7cm} c}
        \hline
        \rowcolor{gray!5}
        \textbf{Variant} & \textbf{Params} & \textbf{\% of full FT} & \textbf{Accuracy (\%)} & \textbf{$\Delta$ (pp)} & \textbf{$95\%$ CI} & \textbf{Sig.} \\ \hline
        Linear probe (no adaptation) & 2{,}052 & 0.11 & $70.40 \pm 2.92$ & --- & --- & --- \\
        FC-LoRA                      & 67{,}588 & 3.67 & $84.58 \pm 1.55$ & $+14.16$ & $[+12.68, +15.64]$ & \checkmark \\
        Conv-LoRA                    & 69{,}636 & 3.78 & $89.16 \pm 1.33$ & $+18.74$ & $[+17.20, +20.25]$ & \checkmark \\
        Conv+FC-LoRA                 & 135{,}172 & 7.34 & $89.84 \pm 2.14$ & $+19.42$ & $[+17.82, +20.97]$ & \checkmark \\ \hline
        \rowcolor{gray!12}
        FC basis-transfer (learned)  & 34{,}820 & 1.89 & $77.29 \pm 2.50$ & $+6.89$ & $[+5.64, +8.10]$ & \checkmark \\
        \rowcolor{gray!12}
        FC random-basis (control)    & 34{,}820 & 1.89 & $73.83 \pm 3.41$ & $+3.43$ & $[+2.62, +4.24]$ & \checkmark \\ \hline
        \rowcolor{gray!12}
        \textbf{CONV basis-transfer (learned)} & \textbf{14{,}340} & \textbf{0.78} & $\mathbf{87.29 \pm 0.86}$ & $\mathbf{+16.86}$ & $\mathbf{[+15.36, +18.32]}$ & \checkmark \\
        \rowcolor{gray!12}
        CONV random-basis (control)  & 14{,}340 & 0.78 & $81.18 \pm 0.89$ & $+10.77$ & $[+9.53, +12.02]$ & \checkmark \\ \hline
        Full fine-tune \textit{(forgets; see Table~\ref{tab:forgetting})} & 1{,}841{,}424 & 100.00 & $94.11 \pm 0.65$ & $+23.71$ & $[+22.12, +25.36]$ & \checkmark \\ \hline
    \end{tabular}
    
    \vspace{1mm}
    \footnotesize
    \textbf{Learned versus random basis at matched budget:} convolutional $+6.11$ pp (stable across seeds; $\sigma = 0.86$ and $0.89$); decision-layer $+3.46$ pp (not reliable across seeds; $\sigma = 2.50$ and $3.41$).
\end{table*}

\begin{figure}[!t]
\centering
\includegraphics[width=\linewidth]{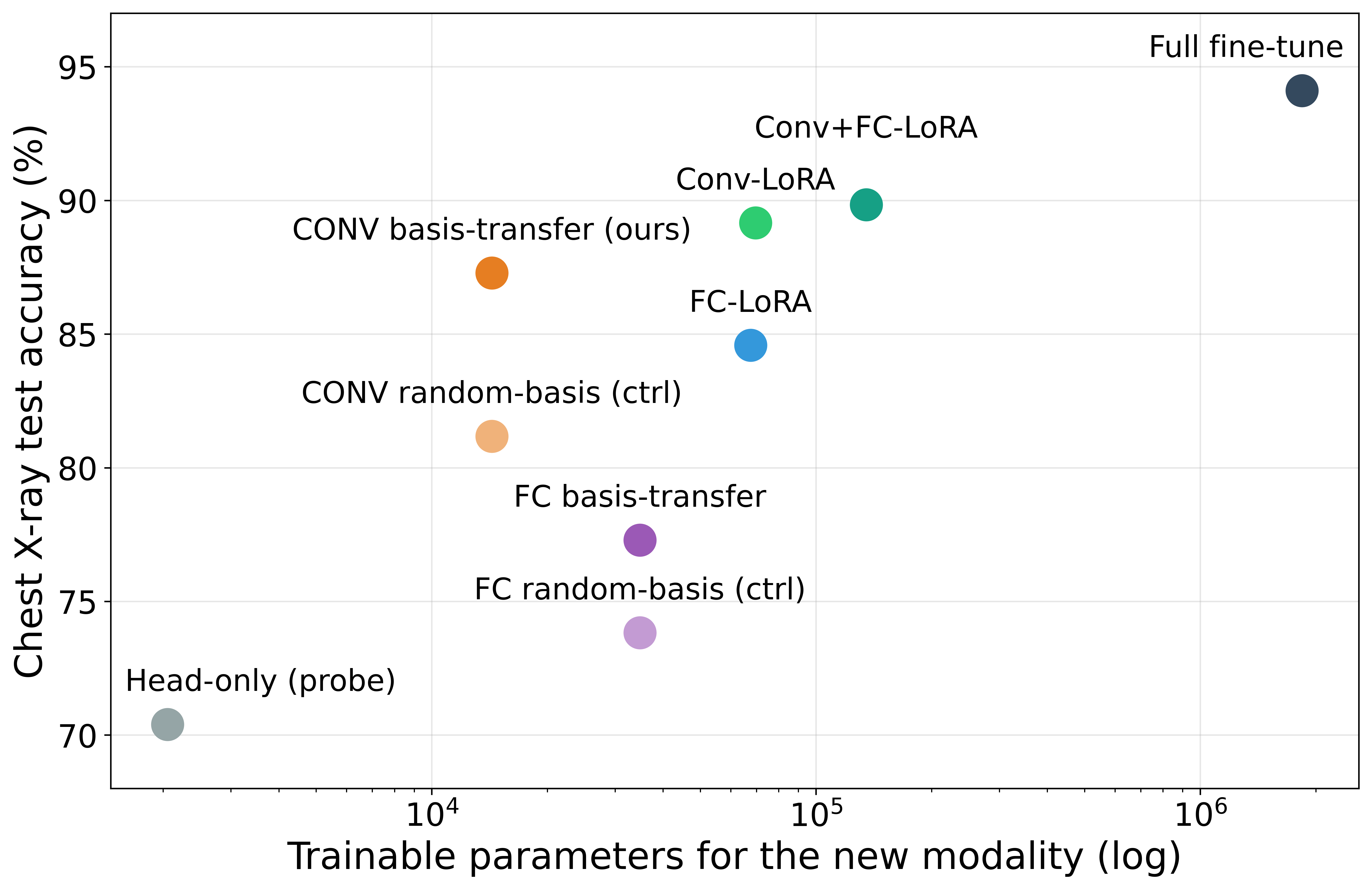}
\caption{Onboarding an unseen modality: accuracy against trainable parameter cost (log scale). Convolutional adaptation dominates decision-layer adaptation at comparable cost. CONV basis-transfer (lower left) attains $87.29\%$ with $14{,}340$ parameters, $0.78\%$ of full fine-tuning.}
\label{fig:result1}
\end{figure}

\subsection{DECISION-LAYER ADAPTATION IS INSUFFICIENT}
\label{subsec:onboarding}

Table~\ref{tab:onboarding} reports all nine variants on the withheld Chest X-ray modality, and Fig.~\ref{fig:result1} plots accuracy against parameter cost.

The linear probe reaches $70.40 \pm 2.92\%$. Because the backbone never observed a radiograph, its frozen features are not adapted to this modality, and a classifier over them is correspondingly limited. Adding a low-rank update at the decision layer (FC-LoRA) raises accuracy to $84.58 \pm 1.55\%$, a significant gain of $+14.16$ pp (CI $[+12.68, +15.64]$), but it remains well short of what is achievable: FC-LoRA can only reweight features the backbone already produces, and cannot recover information the convolutional stages failed to extract.

Adapting the convolutional stages closes most of that shortfall. Conv-LoRA reaches $89.16 \pm 1.33\%$ ($+18.74$ pp over the probe, CI $[+17.20, +20.25]$) at essentially the same parameter cost as FC-LoRA ($69{,}636$ versus $67{,}588$). Combining both (Conv+FC-LoRA, $89.84 \pm 2.14\%$) adds a further $0.68$ pp for twice the parameters, an increment that lies within the seed-to-seed variation and which we therefore do not treat as a meaningful improvement.

The comparison is unambiguous at matched cost: \emph{where} the low-rank update is applied matters far more than how many parameters it contains. This directly answers the question of whether adapter machinery contributes anything beyond a domain-specific linear head. Under all-domains pre-training it does not, because the frozen features are already near-separable. Under the realistic onboarding protocol it contributes $18.74$ pp, but only when the adaptation reaches the convolutional features.

\begin{figure}[!t]
\centering
\includegraphics[width=\linewidth]{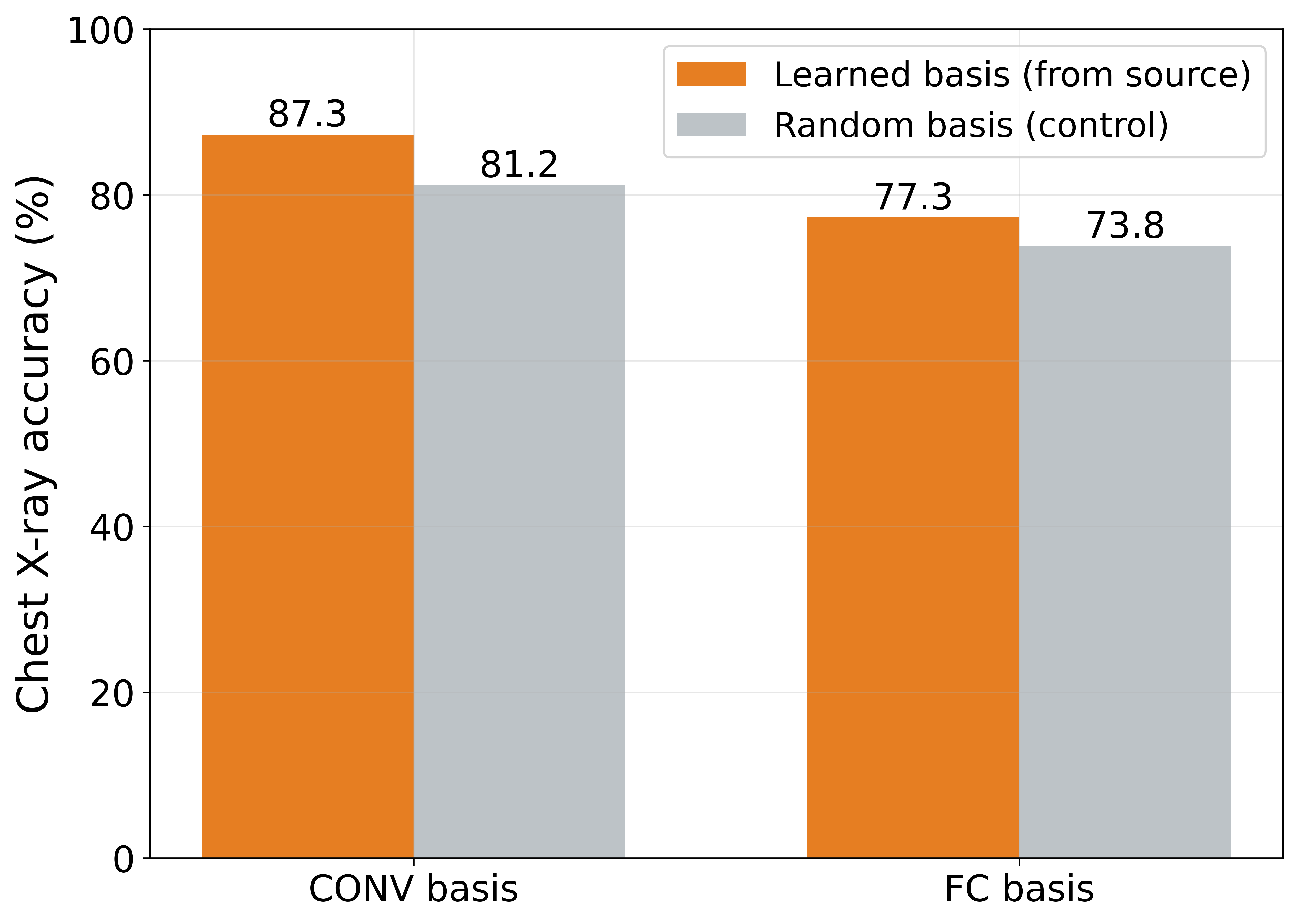}
\caption{Does the learned basis transfer? Each learned basis is compared against a frozen \emph{random} basis with an identical number of trainable parameters. At the convolutional stage the learned basis outperforms the random control by $6.11$ pp; at the decision layer the $3.46$ pp difference is not reliable across seeds.}
\label{fig:result2}
\end{figure}

\subsection{THE CONVOLUTIONAL BASIS TRANSFERS; THE DECISION-LAYER BASIS DOES NOT}
\label{subsec:transfer}

The central question is whether the low-rank basis learned on Kidney CT and Brain MRI carries structure that is useful for a modality it has never encountered. Fig.~\ref{fig:result2} presents the two matched comparisons.

\textbf{Convolutional basis.} Freezing the convolutional basis learned on the source modalities and training only the up-projections attains $87.29 \pm 0.86\%$ with $14{,}340$ trainable parameters. The matched random-basis control, identical in every respect except that its basis is random noise, attains $81.18 \pm 0.89\%$. The learned basis is therefore worth $+6.11$ pp at identical cost. The effect is large relative to its uncertainty (seed standard deviations of $0.86$ and $0.89$ respectively), and it is stable across all three seeds. The convolutional adaptation directions learned from CT and MRI are, in part, modality-agnostic.

\textbf{Decision-layer basis.} The same experiment at the decision layer yields $77.29 \pm 2.50\%$ for the learned basis against $73.83 \pm 3.41\%$ for the random control, a difference of $+3.46$ pp. Although the pooled per-sample bootstrap marks this as significant, the seed-level standard deviations ($2.50$ and $3.41$) are large relative to the difference, and we regard the effect as weak and not reliably established. We report both facts rather than selecting the more favourable one: pooled bootstrap over samples underestimates uncertainty because it does not account for seed-to-seed variation, and where the two disagree the conservative reading should be preferred.

The asymmetry between the two rows is the finding. Transferable low-rank structure exists in medical imaging adaptation, but it resides in the convolutional feature transformation, not in the classifier. This is consistent with the interpretation that early and mid-level convolutional adaptation captures generic properties of grayscale medical images, such as contrast renormalization and edge and texture statistics, which recur across acquisition modalities, whereas decision-layer adaptation is inherently tied to a modality's specific class structure.

The practical consequence is a substantial efficiency gain. CONV basis-transfer recovers $71.2\%$ of the gap between the linear probe and full fine-tuning while training $0.78\%$ of the parameters that full fine-tuning updates, and it comes within $1.87$ pp of a complete convolutional adapter that costs nearly five times as much.

\begin{table*}[t]
    \centering
    \caption{Source-modality accuracy (\%) before and after onboarding the unseen modality. Adapter-based onboarding writes to no shared parameter, so source accuracy is unchanged exactly. Full fine-tuning attains the highest target accuracy in Table~\ref{tab:onboarding} but degrades both source modalities catastrophically.}
    \label{tab:forgetting}
    \small
    \setlength{\tabcolsep}{6pt}
    \begin{tabular}{p{3.2cm} r r r r r}
        \hline
        \rowcolor{gray!5}
        \textbf{Source modality} & \textbf{Before} & \textbf{After (adapter)} & \textbf{$\Delta$ adapter} & \textbf{After (full FT)} & \textbf{$\Delta$ full FT} \\ \hline
        Kidney CT  & 99.05 & \textbf{99.05} & $\mathbf{+0.00}$ & 26.67 & $-72.38$ \\
        Brain MRI  & 90.93 & \textbf{90.93} & $\mathbf{+0.00}$ & 32.53 & $-58.40$ \\ \hline
        \rowcolor{gray!5}
        \textbf{Mean} & \textbf{94.99} & \textbf{94.99} & $\mathbf{+0.00}$ & \textbf{29.60} & $\mathbf{-65.39}$ \\ \hline
    \end{tabular}
\end{table*}

\begin{figure}[!t]
\centering
\includegraphics[width=\linewidth]{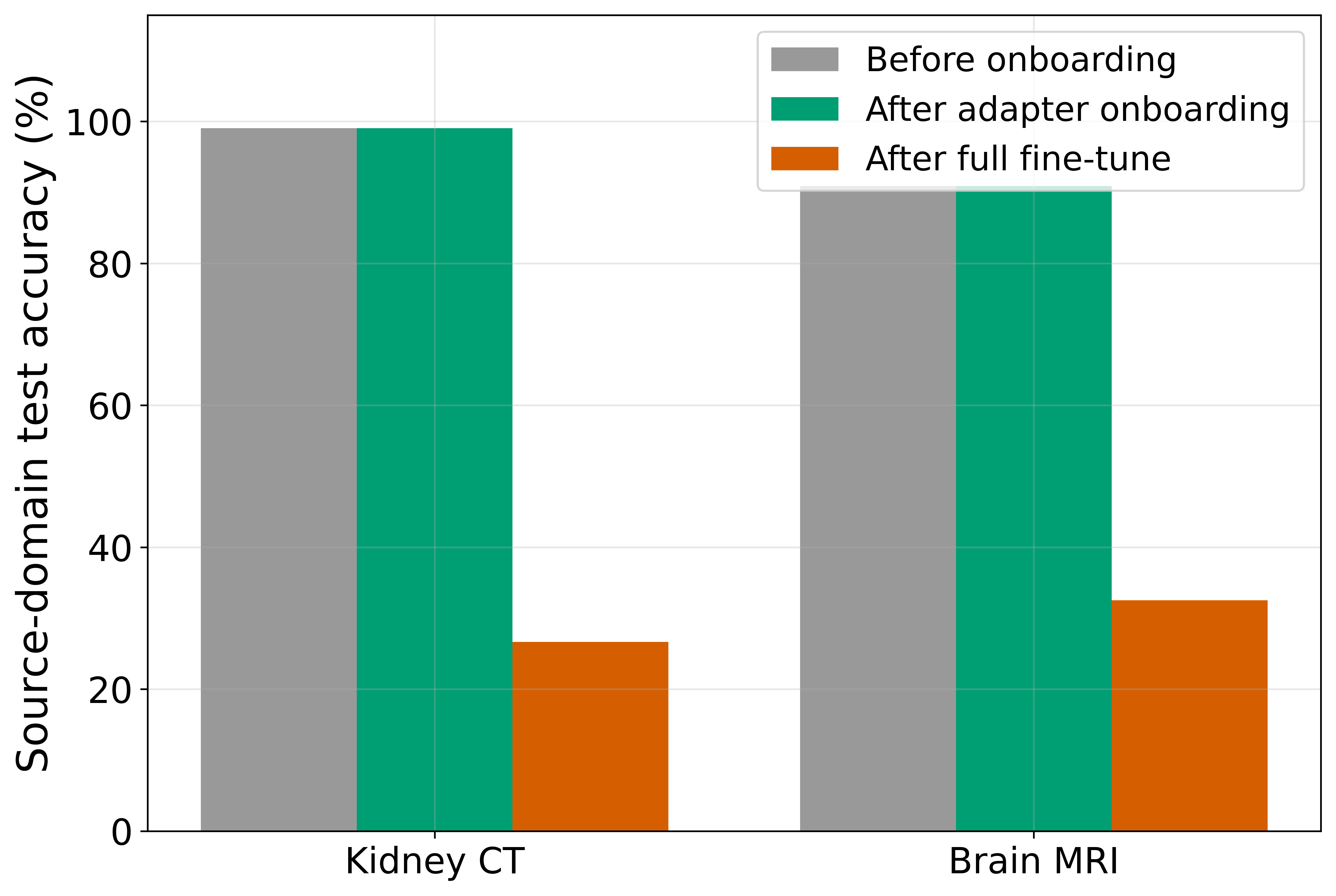}
\caption{Forgetting after onboarding. Adapter-based onboarding leaves source-modality accuracy exactly unchanged, because no shared parameter is written to. Full fine-tuning attains the highest target accuracy but degrades the source modalities by $65.39$ pp on average.}
\label{fig:result3}
\end{figure}

\subsection{ZERO FORGETTING, MEASURED RATHER THAN ASSERTED}
\label{subsec:forgetting}

Adapter-based methods avoid forgetting by construction, and it would be misleading to present this as an algorithmic achievement. It is nonetheless a claim that must be verified, and it acquires meaning only against a baseline that genuinely fails.

Table~\ref{tab:forgetting} and Fig.~\ref{fig:result3} report source-modality accuracy before and after onboarding. Under adapter-based onboarding, Kidney CT remains at $99.05\%$ and Brain MRI at $90.93\%$, unchanged to the last reported digit ($\Delta = 0.00$ pp for both). This is expected: the backbone is frozen, the shared basis is frozen, and onboarding introduces only new up-projections and a new head.

Full fine-tuning tells the opposite story. It attains the highest target accuracy of the study, $94.11 \pm 0.65\%$, but it pays for it by rewriting the shared representation. Kidney CT collapses from $99.05\%$ to $26.67\%$ ($-72.38$ pp) and Brain MRI from $90.93\%$ to $32.53\%$ ($-58.40$ pp), an average degradation of $65.39$ pp. In a deployed system this is not a trade-off but a failure: the modalities already in clinical service would cease to function.

The honest summary is therefore that full fine-tuning holds a $4.27$ pp advantage over the best adapter configuration on the new modality, and that this advantage is unusable. Convolutional adaptation obtains most of the accuracy at a small fraction of the cost while leaving the existing system exactly intact.

\begin{table}[t]
    \centering
    \caption{Convolutional capacity sweep on the unseen modality: rank $r$ against the set of adapted blocks $\mathcal{L}$. ``Gap closed'' is the fraction of the interval between the linear probe and full fine-tuning that is recovered. \textbf{This sweep was run on a single seed}, and the response to rank is non-monotonic in several rows (noise); it characterises the capacity trend rather than identifying a precise optimum. For the same reason the $\mathcal{L}=\{3,4\},\, r{=}16$ entry ($87.29$) is a single-seed value and differs from the three-seed mean for Conv-LoRA in Table~\ref{tab:onboarding} ($89.16 \pm 1.33$).}
    \label{tab:conv_sweep}
    \small
    \setlength{\tabcolsep}{3.5pt}
    \begin{tabular}{l r r r r r}
        \hline
        \rowcolor{gray!5}
        \textbf{Blocks $\mathcal{L}$} & \textbf{$r$} & \textbf{Params} & \textbf{Acc. (\%)} & \textbf{Gap closed (\%)} & \textbf{\% full FT} \\ \hline
        $\{4\}$        & 8   & 24{,}580  & 84.58 & 54.4 & 1.33 \\
        $\{4\}$        & 16  & 47{,}108  & 85.33 & 58.3 & 2.56 \\
        $\{4\}$        & 32  & 92{,}164  & 85.70 & 60.3 & 5.01 \\
        $\{4\}$        & 64  & 182{,}276 & 84.02 & 51.5 & 9.90 \\ \hline
        $\{3,4\}$      & 8   & 35{,}844  & 80.28 & 31.9 & 1.95 \\
        $\{3,4\}$      & 16  & 69{,}636  & 87.29 & 68.6 & 3.78 \\
        $\{3,4\}$      & 32  & 137{,}220 & 84.30 & 52.9 & 7.45 \\
        $\{3,4\}$      & 64  & 272{,}388 & 82.99 & 46.1 & 14.79 \\ \hline
        $\{2,3,4\}$    & 8   & 41{,}476  & 87.94 & 72.1 & \textbf{2.25} \\
        $\{2,3,4\}$    & 16  & 80{,}900  & 88.50 & 75.0 & 4.39 \\
        $\{2,3,4\}$    & 32  & 159{,}748 & 88.97 & 77.5 & 8.68 \\
        $\{2,3,4\}$    & 64  & 317{,}444 & \textbf{89.35} & \textbf{79.4} & 17.24 \\ \hline
        $\{0,\dots,4\}$ & 8  & 44{,}764  & 85.98 & 61.8 & 2.43 \\
        $\{0,\dots,4\}$ & 16 & 87{,}476  & 87.94 & 72.1 & 4.75 \\
        $\{0,\dots,4\}$ & 32 & 172{,}900 & 85.42 & 58.8 & 9.39 \\
        $\{0,\dots,4\}$ & 64 & 343{,}748 & 86.26 & 63.2 & 18.67 \\ \hline
    \end{tabular}
\end{table}

\begin{figure}[!t]
\centering
\includegraphics[width=\linewidth]{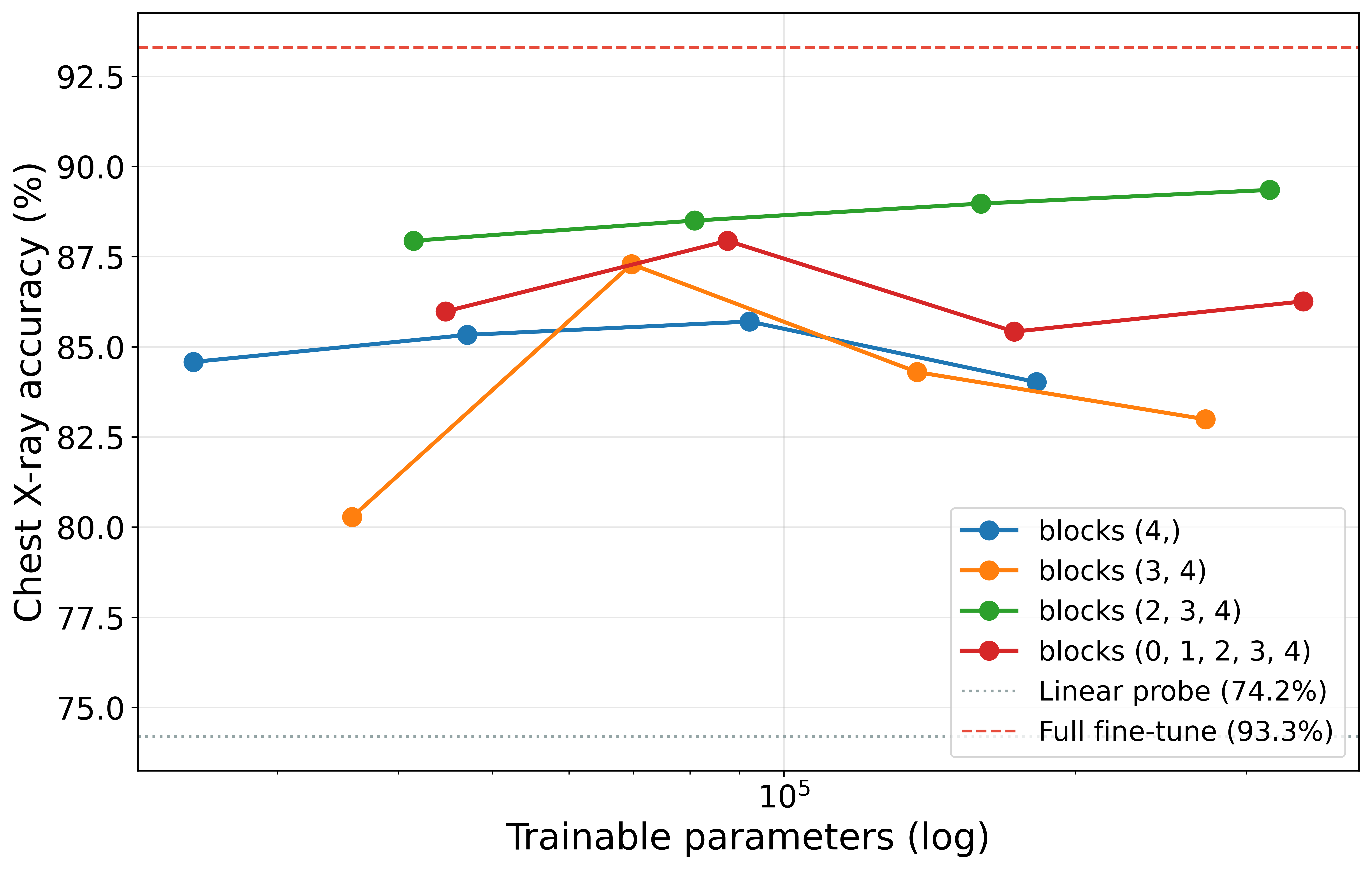}
\caption{Convolutional capacity frontier. Adapting more blocks helps; increasing rank within a fixed block set does not reliably help. The remaining gap to full fine-tuning is not closed by additional convolutional capacity.}
\label{fig:result4}
\end{figure}

\subsection{HOW MUCH CAN CONVOLUTIONAL CAPACITY CLOSE THE GAP?}
\label{subsec:capacity}

Since convolutional adaptation is what works, we ask how far it can be pushed. Table~\ref{tab:conv_sweep} and Fig.~\ref{fig:result4} sweep the rank $r \in \{8,16,32,64\}$ against the set of adapted blocks $\mathcal{L}$.

Two observations follow. First, \emph{block coverage matters more than rank}. Extending adaptation from the last block to the last three ($\mathcal{L} = \{2,3,4\}$) improves accuracy consistently, with the best configuration reaching $89.35\%$ at $r=64$, closing $79.4\%$ of the probe-to-full-fine-tuning gap using $17.2\%$ of full fine-tuning's parameters. Notably, the same block set at $r=8$ already attains $87.94\%$ using only $2.25\%$ of those parameters, an efficient operating point. Adapting all five blocks does not improve further, suggesting that the earliest layers, which encode generic low-level filters, require little modality-specific correction.

Second, and importantly, \emph{additional convolutional capacity does not close the remaining gap to full fine-tuning}. The best swept configuration ($89.35\%$) barely exceeds the default Conv-LoRA setting ($89.16\%$), and both remain roughly five points below full fine-tuning. The residual advantage of full fine-tuning evidently derives from modifying the frozen convolutional filters themselves, which no additive low-rank correction over a frozen basis can replicate. We state this plainly rather than presenting the sweep as though it approached the upper bound.

We also caution that this sweep was run on a single seed, and the response to rank is non-monotonic in several rows: within $\mathcal{L}=\{4\}$, rank $64$ ($84.02\%$) underperforms rank $32$ ($85.70\%$), and within $\mathcal{L}=\{3,4\}$, rank $8$ produces an outlying $80.28\%$. These non-monotonicities are attributable to seed noise. The sweep should be read as a coarse characterization of the capacity trend, not as a precise identification of an optimum.

\begin{table}[t]
    \centering
    \caption{Detecting the unseen modality with a router trained on the source modalities only. AUROC is reported for completeness, but conclusions are drawn from the detection rate at the threshold admitting $95\%$ of source inputs, which is the operating point at which such a gate would actually be used. The two criteria disagree sharply: max-softmax appears adequate by AUROC yet misses a third of the unseen modality in practice.}
    \label{tab:ood}
    \small
    \setlength{\tabcolsep}{6pt}
    \begin{tabular}{p{3.4cm} r r}
        \hline
        \rowcolor{gray!5}
        \textbf{Score} & \textbf{AUROC} & \textbf{Detection @ $95\%$ TNR} \\ \hline
        Max-softmax \cite{Hendrycks2017MSP}   & 0.9008 & 65.8\% \\
        Energy \cite{Liu2020Energy}           & 0.8741 & 57.3\% \\
        \textbf{Mahalanobis} \cite{Lee2018Mahalanobis} & \textbf{0.9998} & \textbf{100.0\%} \\ \hline
    \end{tabular}
\end{table}

\begin{figure}[!t]
\centering
\includegraphics[width=\linewidth]{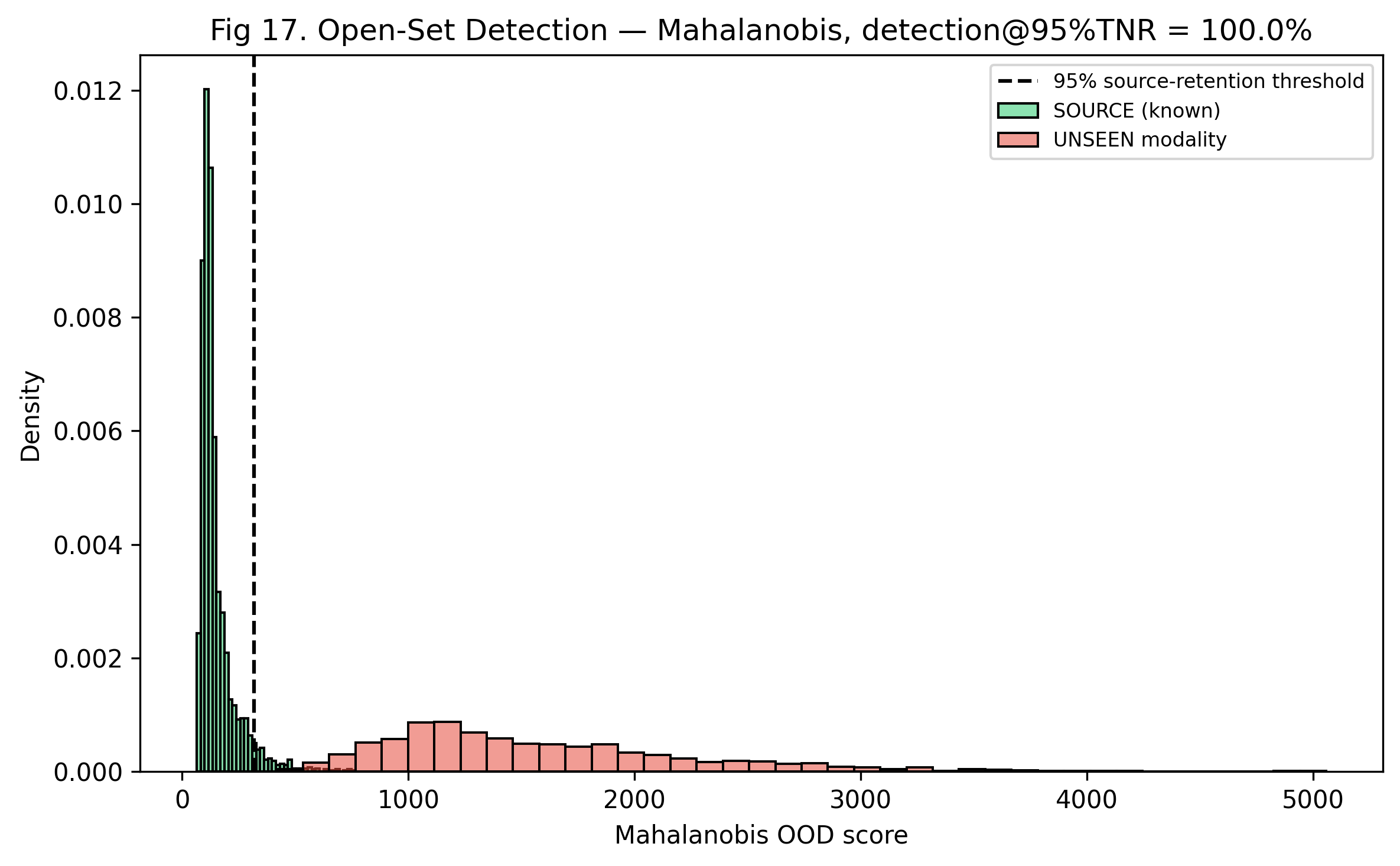}
\caption{Open-set detection of the unseen modality using the best-performing score. The Mahalanobis distance on frozen backbone features separates the withheld modality from the source modalities almost perfectly.}
\label{fig:result5}
\end{figure}

\subsection{DETECTING WHEN ONBOARDING IS REQUIRED}
\label{subsec:ood_results}

Onboarding is actionable only if the arrival of an unseen modality can be detected. Table~\ref{tab:ood} evaluates the three scores of Eqs.~\ref{eq:msp}--\ref{eq:maha}, and Fig.~\ref{fig:result5} shows the score distributions for the best detector.

The three scores differ sharply, and the choice of metric determines whether one would draw the right conclusion. Max-softmax attains an AUROC of $0.9008$, which appears acceptable, yet flags only $65.8\%$ of chest radiographs at the threshold admitting $95\%$ of source inputs. The energy score is worse on both counts (AUROC $0.8741$, detection $57.3\%$). A router trained to discriminate source modalities is confidently wrong on inputs from outside its repertoire, which is a well-documented failure of softmax-based confidence and a reason we did not rely on it.

The Mahalanobis score on frozen backbone features behaves entirely differently, attaining an AUROC of $0.9998$ and detecting $100\%$ of the unseen modality at $95\%$ source retention. Because it measures distance to the source feature distribution rather than confidence within a closed label set, it is not misled by the router's overconfidence. This provides a practical trigger: a system can reliably determine that an input lies outside its current modality repertoire, and therefore that onboarding, rather than prediction, is the appropriate response.

We stress the methodological point. Had we judged these detectors by AUROC alone, max-softmax ($0.90$) would have appeared adequate; at the operating point that matters it misses one in three cases. Deployment claims about safety mechanisms should be evaluated at the threshold at which they would actually be used.

\begin{table}[t]
    \centering
    \caption{Parameter accounting, separating the three quantities that are commonly conflated. The backbone and the shared convolutional basis are stored once and reused by every modality; only the up-projections and the classification head are paid per newly onboarded modality. The marginal cost of a new modality does not grow with the number of modalities already supported.}
    \label{tab:params}
    \small
    \setlength{\tabcolsep}{4pt}
    \begin{tabular}{p{4.0cm} r p{2.4cm}}
        \hline
        \rowcolor{gray!5}
        \textbf{Component} & \textbf{Params} & \textbf{Cost model} \\ \hline
        Frozen backbone                     & 1{,}839{,}372 & Stored once \\
        Shared conv.\ basis $\mathcal{A}$ ($\mathcal{L}{=}\{3,4\}$, $r{=}16$) & 55{,}296 & Stored once \\ \hline
        Up-projections $\mathcal{B}^{(\text{tgt})}$ & 12{,}288 & Per modality \\
        Target head                          & 2{,}052   & Per modality \\ \hline
        \rowcolor{gray!12}
        \textbf{Marginal cost per new modality} & \textbf{14{,}340} & \textbf{0.78\% of full FT} \\ \hline
        Full fine-tuning (per modality)      & 1{,}841{,}424 & 100\% \\
        Separate full model (per modality)   & 1{,}839{,}372 & 99.9\% \\ \hline
    \end{tabular}
\end{table}

\subsection{PARAMETER ACCOUNTING}
\label{subsec:params}

Table~\ref{tab:params} separates three quantities that are frequently conflated: total stored parameters, trainable parameters per newly onboarded modality, and inference-time active parameters. The frozen backbone ($1.84$M) is stored once and shared by all modalities; the shared convolutional basis ($55{,}296$ parameters for $\mathcal{L}=\{3,4\}$, $r=16$) is likewise stored once. Onboarding an additional modality by basis transfer costs only $14{,}340$ new parameters, of which $12{,}288$ are up-projections and $2{,}052$ the classification head.

Consequently the marginal cost of a new modality is $0.78\%$ of full fine-tuning, and this cost does not grow with the number of modalities already supported. Maintaining separate full models, by contrast, would cost $1.84$M parameters per modality.

\subsection{LIMITATIONS}
\label{subsec:limitations}

Several limitations bound the scope of these conclusions.

\textit{Single transfer direction.} We establish that a basis learned on CT and MRI transfers to X-ray. We have not tested the reverse directions, nor characterized when transfer succeeds or fails. Whether this reflects a general property of medical imaging adaptation or a favourable property of this particular source-target combination remains open, and the full leave-one-out matrix over more modalities is the natural next experiment.

\textit{Architecture and scale.} The backbone is a $1.84$M-parameter convolutional network on $128 \times 128$ inputs. Whether convolutional basis transfer extends to attention-based backbones, or to large-scale pre-trained vision models, is untested and cannot be assumed.

\textit{Patient-level splitting.} The public datasets used here do not expose patient identifiers, so patient-disjoint splits cannot be guaranteed. Reported accuracies may therefore partly reflect dataset-specific acquisition signatures rather than purely disease-level features. This affects all variants equally and does not confound the \emph{relative} comparisons, which are the basis of our claims, but it does limit any absolute clinical interpretation.

\textit{Capacity sweep resolution.} The convolutional capacity sweep was run on a single seed and exhibits noise-level non-monotonicity, as noted in Section~\ref{subsec:capacity}.

\textit{Mechanism.} We demonstrate that the learned convolutional basis transfers, and we quantify by how much, but we do not explain what it encodes. A subspace analysis of the learned versus random bases would convert this empirical observation into a mechanistic account, and we regard it as the most valuable follow-up.

\section{CONCLUSION}
\label{sec:Conclusion}

We studied the problem of onboarding an imaging modality that a deployed, frozen backbone has never seen, under a leave-one-domain-out protocol in which the target modality is withheld from pre-training entirely. This protocol matters: when a backbone is pre-trained on all of the domains it will later adapt to, its frozen features are already near-separable for each of them, a per-domain linear head suffices, and the resulting evaluation says nothing about the deployment scenario that motivates parameter-efficient adaptation in the first place.

Under the realistic protocol, three conclusions follow. First, decision-layer adaptation is insufficient: a linear probe and fully connected LoRA both fall well short on the unseen modality, whereas convolutional LoRA recovers most of the achievable accuracy at comparable parameter cost, so what is adapted matters more than how much is adapted. Second, the low-rank convolutional basis learned on the source modalities transfers: freezing it and training only the up-projections onboards the unseen modality at $0.78\%$ of the cost of full fine-tuning and exceeds an identically sized random basis by $6.11$ pp, while the corresponding decision-layer basis shows no reliable transfer, localizing the transferable structure to the convolutional stage. Third, adapter-based onboarding leaves source-modality accuracy exactly unchanged ($\Delta = 0.00$ pp), whereas full fine-tuning attains the highest target accuracy only by degrading the source modalities catastrophically, a trade that is unusable in a deployed system. A Mahalanobis score on frozen features detects the unseen modality reliably, providing a practical trigger for when onboarding should occur.

Taken together, these results describe a workable procedure for extending a medical imaging system to a new modality: detect that the modality is out of repertoire, reuse the convolutional adaptation basis already learned from the modalities in service, and train only a small set of up-projections and a classifier head. The existing system is left bit-for-bit intact.

Several directions follow from the limitations of this study. The most immediate is to establish the \emph{generality} of convolutional basis transfer by evaluating the full leave-one-out matrix over a larger set of modalities, including dermoscopy, retinal fundus imaging, and histopathology, and by characterizing which source combinations transfer to which targets; a predictive account, forecasting from source-domain statistics alone whether a basis will transfer to a given target, would be more valuable still. The mechanism also deserves explanation: a subspace analysis comparing the learned and random bases, for instance through their principal angles or the filters they induce, would identify what the transferable directions encode and would turn the present empirical result into a mechanistic one. Whether an analogous basis exists in the attention projections of vision transformers, or in large-scale pre-trained vision backbones, would in turn determine whether this is a property of convolutional adaptation specifically or of low-rank medical-imaging adaptation in general.

Clinical validation, finally, requires data that these public benchmarks cannot provide: patient-level disjoint splits, multi-institutional evaluation, and analysis of scanner and protocol variation within a single modality. Establishing that the reported gains survive such conditions is a prerequisite to any deployment claim.


\bibliographystyle{IEEEtran}  
\bibliography{bibliography/references}



\begin{IEEEbiography}[{\includegraphics[width=1in,height=1.25in,clip,keepaspectratio]{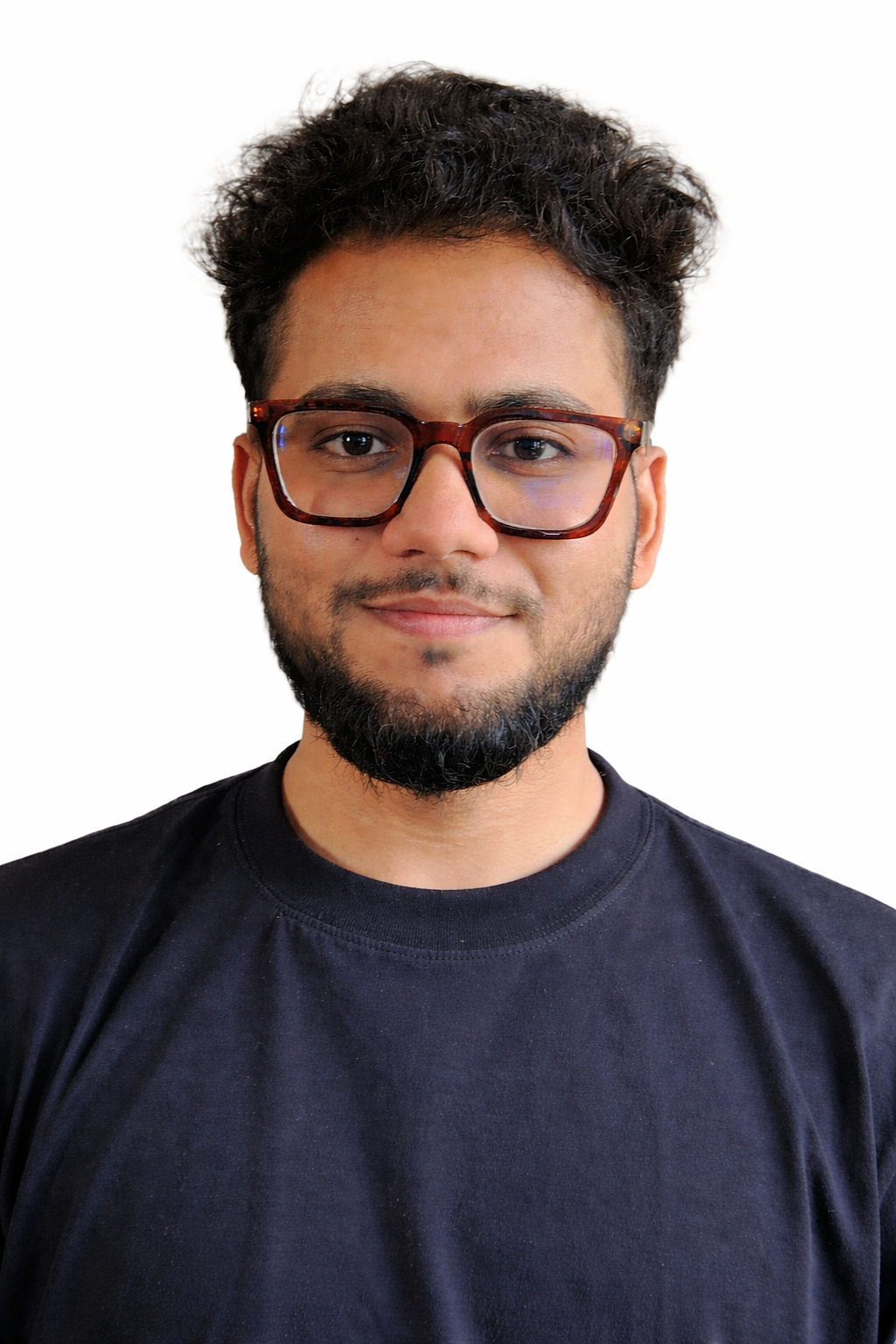}}]{Ranat Das Prangon} ~is an undergraduate student in Chemical Engineering at the Bangladesh University of Engineering and Technology. His research interests focus on neural networks, deep learning, and their applications in computer vision and intelligent systems. His work involves designing and optimizing deep learning models for real-world applications, including classification and pattern recognition tasks. In addition to research, he serves as the Head of Product Innovation at HBitCorp, contributing to AI-driven systems, and as the Chief Technology Officer of ACS Doubts, where he leads the development of scalable AI-based educational platforms.

\end{IEEEbiography}

\vspace{-2.0em}   

\begin{IEEEbiography}[{\includegraphics[width=1in,height=1.25in,clip,keepaspectratio]{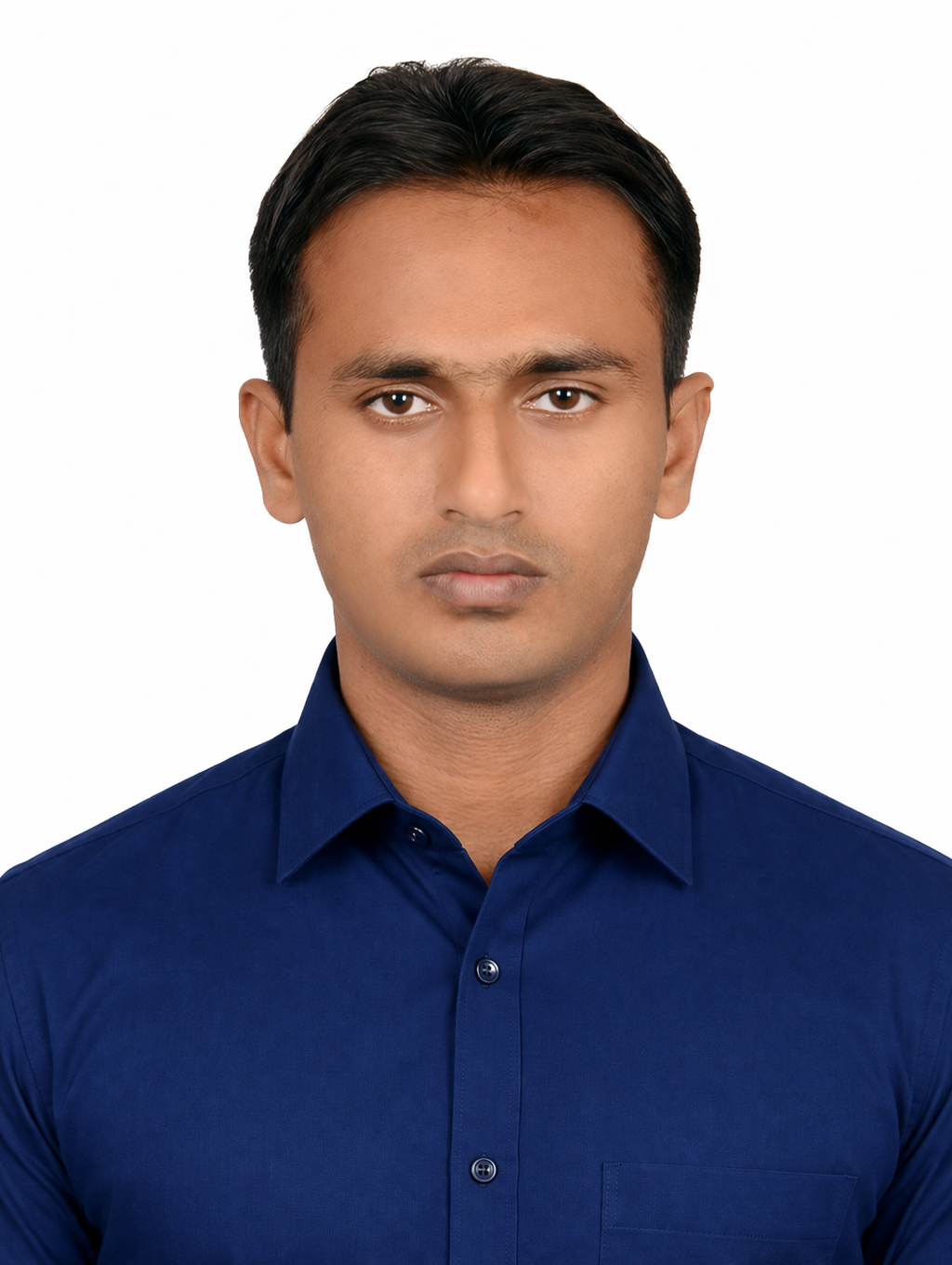}}]{Istiaque Ahmed}  ~received the Ph.D. degree in Informatics from Osaka Metropolitan University, Osaka, Japan, in 2026. He also received the bachelor’s and master’s degrees in Computer Science and Engineering from the University of Rajshahi, Rajshahi, Bangladesh, in 2013 and 2015, respectively. He worked as a Research Scientist with the Blockchain Economy Research Center, GIST, Korea, and has experience as a Senior Software Engineer and Blockchain Solution Consultant. He also served as a Lecturer at multiple Universities in Bangladesh. His research interests include blockchain, SSI, DID, eKYC, ZKPs, privacy and security, Web3.0, DeFi, AI, LLMs, and agentic AI.
\end{IEEEbiography}

\vspace{-2.0em}   

\begin{IEEEbiography}[{\includegraphics[width=1in,height=1.25in,clip,keepaspectratio]{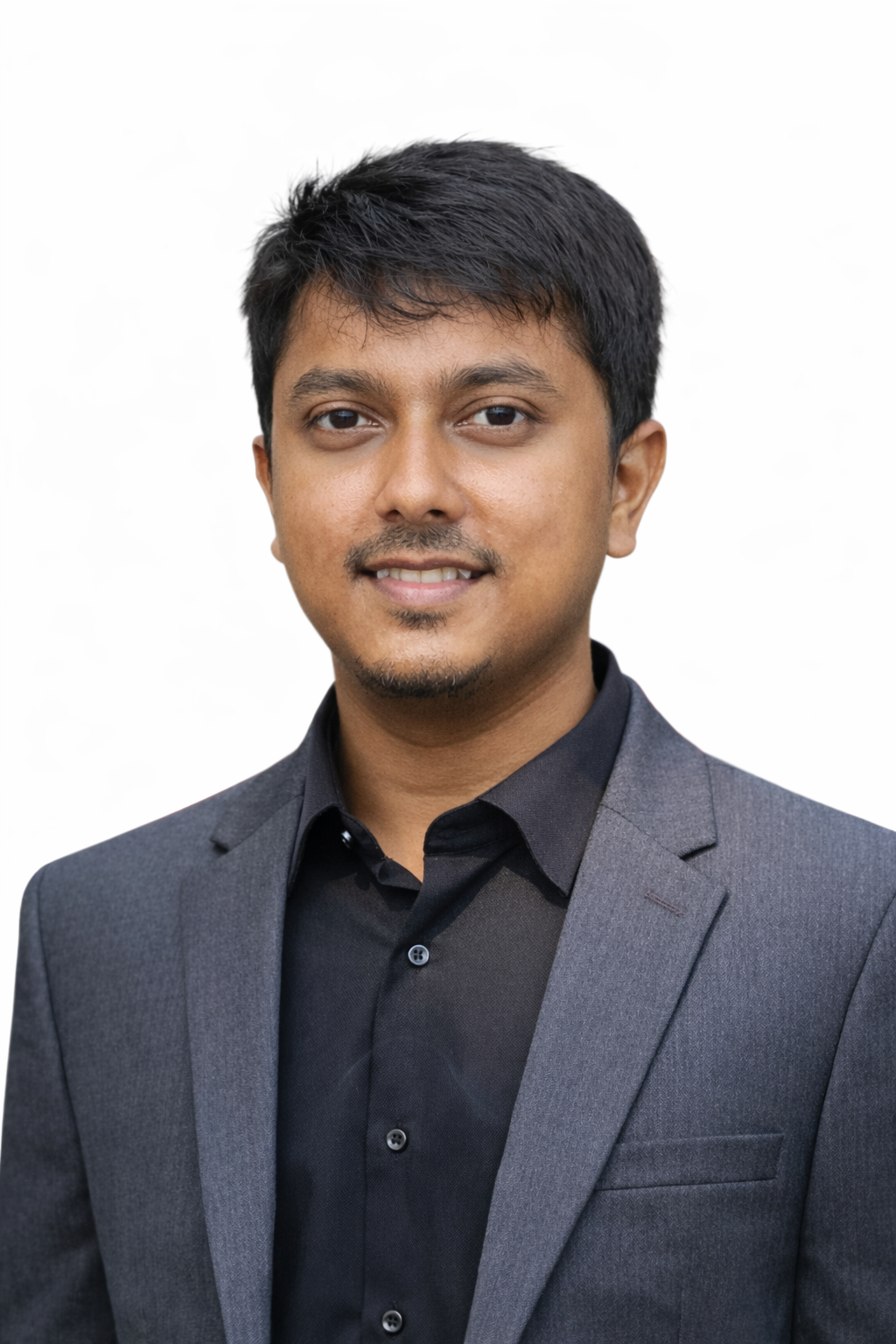}}]{Shajid Hasan Naim} ~is an undergraduate student in Mechanical Engineering at the Chittagong University of Engineering and Technology. His research interests focus on neural networks, physics-informed neural networks, and their applications in fluid mechanics and aerodynamics, with current work on fluid field reconstruction using physics-informed approaches. He has developed impactful software systems, including Chithi.me and an open-source npm package. He is also interested in future research at the intersection of machine learning and physical sciences.
\end{IEEEbiography}

\vspace{-2.0em}   

\begin{IEEEbiography}[{\includegraphics[width=1in,height=1.25in,clip,keepaspectratio]{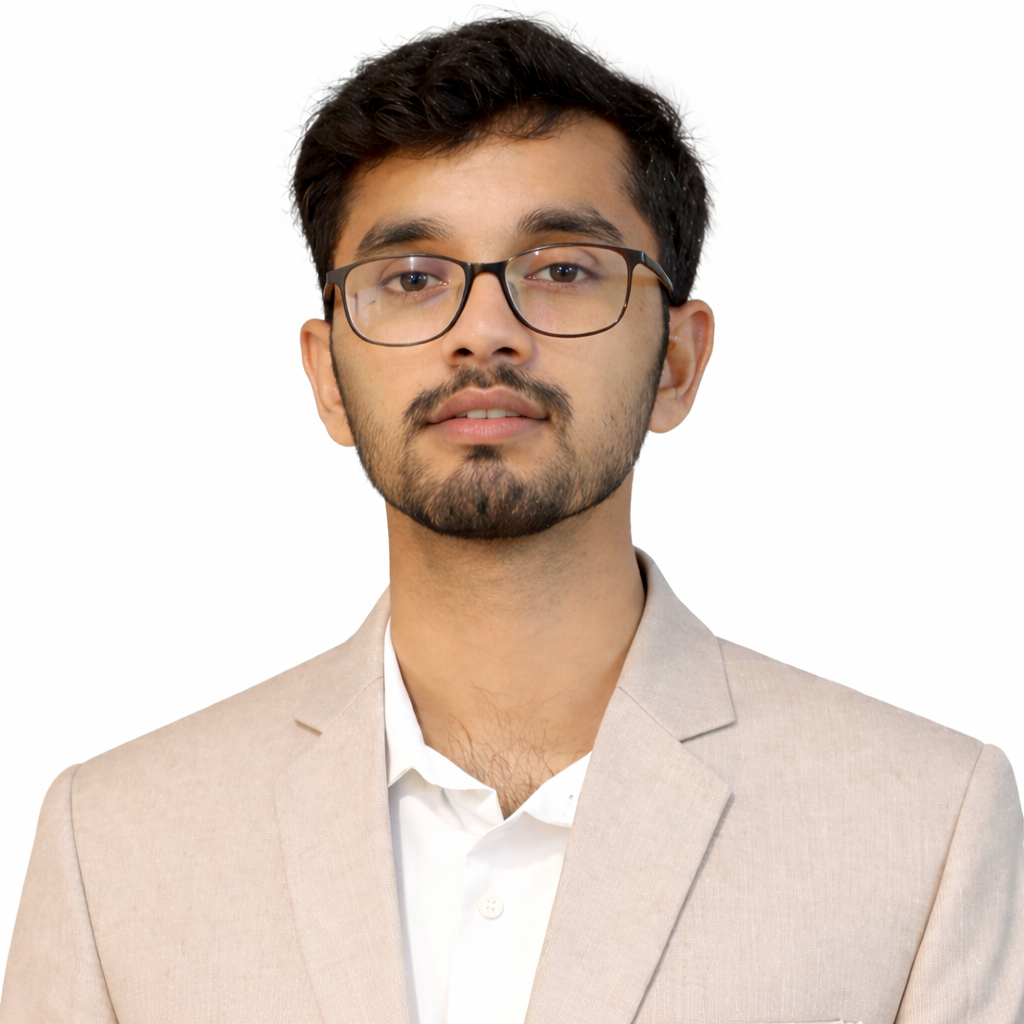}}]{ Waseem Mustak Zisan} ~is currently pursuing a Bachelor’s degree in Computer Science and Engineering at Bangladesh University of Engineering and Technology (BUET), Dhaka, Bangladesh. His research interests include graph theory and machine learning. He has worked as a Software Engineer at Merilsoft LLC, New York, USA, and has experience in backend systems .He has also worked as a Competitive Programming Coach at Manarat International University, Dhaka, Bangladesh, contributing to training students in data structures, algorithms, and problem-solving.
\end{IEEEbiography}

\vspace{-2.0em}   

\begin{IEEEbiography}[{\includegraphics[width=1in,height=1.25in,clip,keepaspectratio]{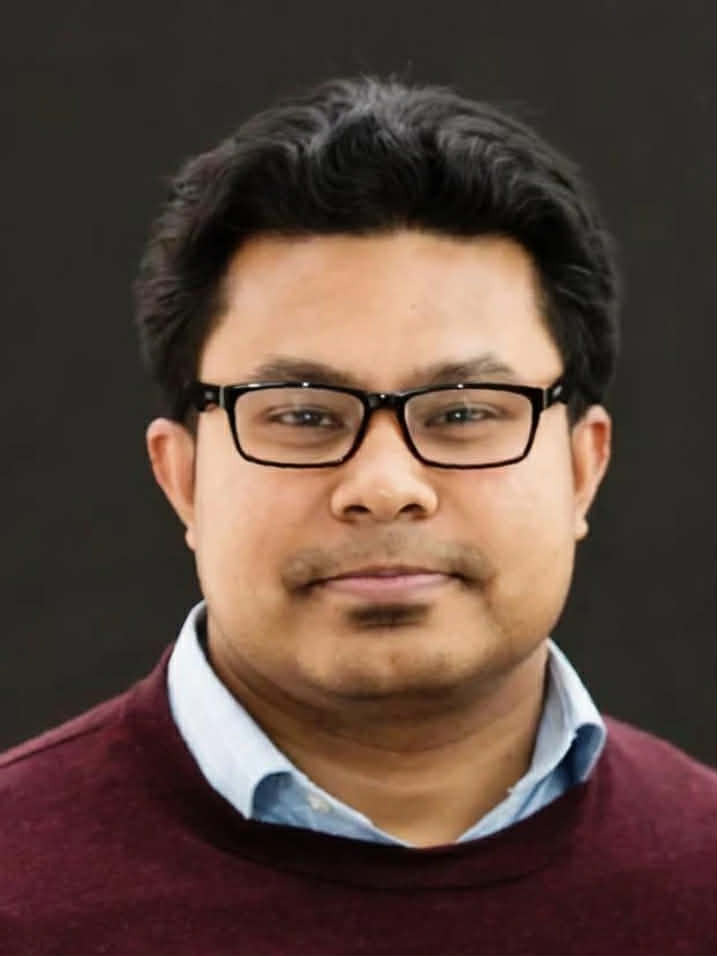}}]{Md Shakhawat Hossain, PhD,}  ~is an Associate Professor at the School of Informatics, Kochi University of Technology (KUT), Japan, where he leads the AI for Medical Imaging (AIM) Laboratory. He received his Ph.D. from the Tokyo Institute of Technology, Japan, in collaboration with Memorial Sloan Kettering Cancer Center (MSK), USA. He later worked as a Research Fellow at MSK and as a Senior Researcher in Machine Learning for Medical Imaging at the University of Oxford, UK. His research focuses on clinically deployable, lightweight, and explainable AI for medical image analysis, including cancer diagnosis and decision-support systems.
\end{IEEEbiography}

\vfill

\end{document}